%% file: acl_latex.tex
\documentclass[11pt]{article}

\usepackage[preprint]{acl}

\usepackage{times}
\usepackage{latexsym}

\usepackage[T1]{fontenc}

\usepackage[utf8]{inputenc}

\usepackage{microtype}

\usepackage{inconsolata}

\usepackage{graphicx}
\usepackage{pifont}
\usepackage{adjustbox}
\usepackage{multirow}
\usepackage{bbding}
\usepackage{threeparttable}
\usepackage{amsmath}
\usepackage{amssymb}
\usepackage{mathtools}
\usepackage{amsthm}
\usepackage{microtype}
\usepackage{graphicx}
\usepackage{subcaption}
\usepackage{booktabs}
\definecolor{deepgreen}{RGB}{6,153,6} 
\definecolor{deepred}{RGB}{254,34,35}    
\definecolor{deepyellow}{HTML}{FFBF00}
\usepackage{enumitem}
\usepackage{bbm}
\usepackage{array}
\usepackage{longtable}
\usepackage{array}
\usepackage[dvipsnames]{xcolor}
\usepackage[most]{tcolorbox}
\tcbuselibrary{breakable,skins}
\usepackage{algorithm}
\usepackage{algorithmic}
\usepackage{xspace}
\definecolor{headergray}{RGB}{235,235,235}
\definecolor{promptblue}{RGB}{225,239,252}
\definecolor{promptblueframe}{RGB}{91,141,189}
\definecolor{promptbluetitle}{RGB}{31,78,121}
\definecolor{promptbody}{RGB}{248,251,255}
\definecolor{JarvisGreen}{RGB}{88,136,43}
\definecolor{JarvisOrange}{RGB}{204,102,0}
\definecolor{JarvisBlue}{RGB}{36,88,180}
\newtcolorbox{prompttable}[3]{%
  enhanced,
  breakable,
  width=\linewidth,
  colback=promptbody,
  colframe=promptblueframe,
  colbacktitle=promptblue,
  coltitle=promptbluetitle,
  fonttitle=\bfseries,
  title={#3},
  title after break={#3\ (continued)},
  arc=1mm,
  outer arc=1mm,
  rounded corners,
  boxrule=0.45pt,
  left=2mm,
  right=2mm,
  top=1.2mm,
  bottom=1.2mm,
  skin first=enhanced,
  skin middle=enhanced,
  skin last=enhanced,
  extras first={rounded corners, arc=1mm, outer arc=1mm},
  extras middle={rounded corners, arc=1mm, outer arc=1mm},
  extras last={rounded corners, arc=1mm, outer arc=1mm},
  before={%
    \par\smallskip
    \refstepcounter{table}%
    \label{#1}%
  },
  after={%
    \par\smallskip
    \noindent{\small\textbf{Table~\thetable:} #2}
    \par\smallskip
  },
  before upper={\small},
}
\newcommand{\promptsection}[1]{%
  \noindent{\color{promptbluetitle}\rule{\linewidth}{0.4pt}}\\%
  \noindent{\small\bfseries\color{promptbluetitle} #1}%
  \vspace{0.35em}%
}
\newcommand*{\imgintitle}[1]{%
\raisebox{-1ex}[0pt][0pt]{
\includegraphics[height=4ex, keepaspectratio]{#1}%
}%
}
\newcommand{\jarvisbench}{JarvisGUI\xspace}
\title{\imgintitle{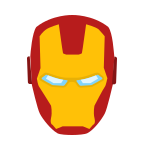}\textsc{\jarvisbench}: Towards Cross-Device GUI Agents \\with Dynamic Task Composition}

\author{
	Zixiang Chen\textsuperscript{1}\thanks{\quad Equal contribution}, 
    Yuheng Lu\textsuperscript{1}\footnotemark[1], Zihao Cheng\textsuperscript{1}, 
    Zeming Liu\textsuperscript{1}\thanks{\quad Corresponding author: Zeming Liu.}, 
    Jizeng Bai\textsuperscript{1}, 
    Ziye Huang\textsuperscript{1},
    \\ \textbf{ 
    Zhiyin Lin\textsuperscript{1}, 
    Zihan Li\textsuperscript{1}, 
    Yuhang Guo\textsuperscript{2},
    Yunhong Wang\textsuperscript{1},
    Haifeng Wang\textsuperscript{3}
    }\\
    \textsuperscript{1}School of Computer Science and Engineering, Beihang University, Beijing, China \\
	\textsuperscript{2}Beijing Institute of Technology, Beijing, China, \quad
	\textsuperscript{3}Baidu Inc., Beijing, China \hspace{1mm}\\
	{\tt \{chenzixiang, luyuh, zmliu\}@buaa.edu.cn} 
    }

\begin{document}
\maketitle
\input{sections/0_abstract}




\input{sections/1_introduction}

\input{sections/2_related_works}
\input{sections/3_environment}
\input{sections/4_JarvisBench}

\input{sections/5_experiment}
\input{sections/6_Conclusion}

\section*{Limitations}


One limitation is our dependency on specific system resources, notably Docker and the KVM module, to run parallel virtual machines. This may hinder reproducibility for researchers in restricted or cloud-based environments lacking KVM support; however, those affected are encouraged to contact us for access to our experimental setup. Furthermore, while \jarvisbench covers diverse everyday workflows, it currently lacks accessibility-focused scenarios for elderly and disabled users, and its scenario coverage is constrained by practical trade-offs involving legal compliance, evaluation costs, and hardware simulation limits. We detail these design choices and identify key directions for future benchmark expansion in Appendix \ref{sec:appendix_future_work}.

\section*{Ethical Statement}
This study ensures data privacy by anonymizing collected tasks and excluding sensitive information. Manual annotations were strictly supervised to minimize bias, and AI tools (DeepSeek, Qwen) were utilized solely for code and language refinement. We properly attribute all open-source components and will release the dataset under agreements that balance transparency with data protection. Finally, we acknowledge the dual-use risks of cross-device GUI agents: their advanced automation capabilities could potentially be exploited for malicious file theft or unauthorized data leakage. Mitigating these threats underscores the need for future research into robust permission boundaries, secure agent alignment, and OS-level anomaly detection.
    
\section*{Acknowledgments}

Thanks for the insightful comments and feedback from the reviewers. This work was supported by the National Natural Science Foundation of China (No. 62406015).

\bibliography{custom}

\appendix
\input{appendix}


\end{document}

%% file: sections/0_abstract.tex
\begin{abstract}
Real-world GUI usage frequently involves workflows that span multiple devices and platforms, requiring the transfer of intermediate results, maintenance of shared state, and coordination across heterogeneous environments. However, existing GUI benchmarks overwhelmingly evaluate agents on single-device, statically defined tasks, thus leaving such cross-device capabilities largely unexamined, resulting in an overly optimistic assessment of agents' readiness for real-world usage. We introduce \jarvisbench, a dynamic benchmark that evaluates GUI agents on cross-device workflows requiring coordinated interaction across heterogeneous platforms—including Android, Windows, and Ubuntu. Specifically, \jarvisbench formulates GUI tasks as input–output transformations under a lightweight type system, which allows us to automatically compose multi-step, cross-device workflows and dynamically evaluate agent performance within a unified framework. By evaluating agents in virtual environments spanning multiple operating systems, \jarvisbench reveals that state-of-the-art open-source GUI agents struggle with the state-transfer awareness, cross-platform contextual reasoning, and long-horizon dependency management required for real-world workflows, exposing a critical capability gap invisible to existing benchmarks \footnote{Our code and data will be publicly available at \href{https://github.com/BUAA-IRIP-LLM/JarvisGUI}{\texttt{JarvisGUI}}.}.
\end{abstract}

%% file: sections/1_introduction.tex
\begin{figure*}[!ht]
    \centering
    \includegraphics[width=1\linewidth]{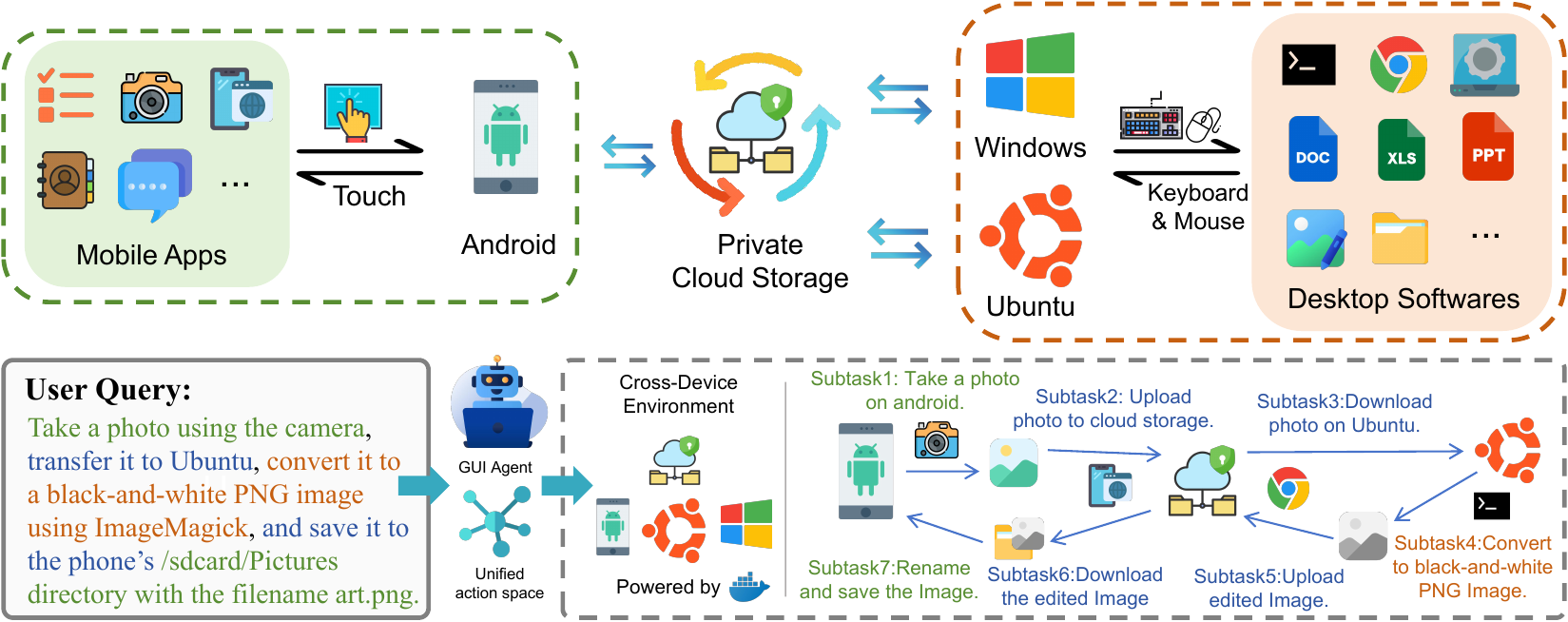}
\caption{Overview of \jarvisbench. {\color{JarvisGreen}Green} denotes tasks on mobile platforms, {\color{JarvisOrange}orange} denotes tasks on desktop platforms, and {\color{JarvisBlue}blue} denotes cross-device transfer tasks.}\label{fig:intro}
\end{figure*}
\section{Introduction}
The rapid advancement of Multimodal Large Language Models (MLLMs)~\citep{yin2024survey, zhao2023survey, xie2024large} has catalyzed the development of Graphical User Interface (GUI) agents that can perceive complex visual environments and autonomously execute tasks through simulated human interactions (e.g., clicking, typing, and scrolling)~\citep{wang2024gui, yin2024survey, nguyen2025gui}. To systematically measure this progress, recent benchmarks have been proposed to evaluate GUI agents in both static~\citep{omniact, seeclick, screenspotpro, guiodyssey} and dynamic environments~\citep{osworld, crab, omnibench,liu2026docosproactivedocumentguidedactions}.

 However, these benchmarks predominantly adopt a \textbf{single-device} assumption, where each task is confined to one operating system or platform. This paradigm neglects a fundamental characteristic of real-world digital workflows: the necessity of \textbf{cross-device} interaction. In practice, users routinely coordinate actions across smartphones, personal computers, and servers~\citep{brudy2018overview,raptis2016continuity,majrashi2021crossdevice,jokela2015diary}. For example, Figure~\ref{fig:intro} shows such a cross-device workflow. Existing benchmarks fail to capture this cross-device dependency, resulting in an incomplete assessment of agent capabilities under realistic deployment conditions.

To bridge this gap, we propose \textbf{\jarvisbench}, a novel benchmark specifically designed to evaluate GUI agents in cross-device settings. Unlike prior benchmarks that focus on isolated environments, \jarvisbench introduces a heterogeneous multi-device ecosystem comprising \texttt{Android}, \texttt{Windows}, and \texttt{Ubuntu} platforms. Tasks in \jarvisbench explicitly require agents to coordinate actions across devices, including transferring information, sharing files, and maintaining task context throughout multi-platform execution.

Using \jarvisbench, we conduct comprehensive evaluations of several representative open-source GUI agents~\citep{qwen3vl, mai-ui, gui-owl, ui-venus}. 
To accommodate the prevalent single-platform interaction paradigm, in which current GUI agents process only one active platform per step, we adopt a two-stage Planner-Grounder architecture, with details in Appendix~\ref{Planner-Model-Details}.
Results reveal substantial disparities between single-device proficiency and cross-device competence, highlighting persistent challenges in context preservation and inter-device reasoning.

Overall, our contributions are summarized as follows:
\begin{itemize}
    \item To the best of our knowledge, we are the first to systematically study \textbf{cross-device GUI agent tasks}, where successful execution requires transferring information, files, and execution context across heterogeneous devices.

    \item We propose a \textbf{slot-typing-based atomic task modeling framework} that explicitly models transferable entities and inter-task dependencies, enabling reliable, extensible, and verifiable composition of complex cross-device tasks.

    \item Building on this framework, we introduce \textbf{\jarvisbench}, a dedicated benchmark spanning \texttt{Android}, \texttt{Windows}, and \texttt{Ubuntu} for evaluating GUI agents in realistic multi-ecosystem scenarios. Experiments with strong open-source agents reveal that cross-device coordination remains a major bottleneck largely obscured by existing GUI benchmarks.

\end{itemize}

%% file: sections/2_related_works.tex
\section{Related Work}

\subsection{GUI Agent}

Recent advances in vision–language models (VLMs) \citep{qwen25vl, qwen3vl} have significantly shaped the development of GUI agents, which aim to execute user instructions through interactions with GUI. Early GUI agents predominantly relied on structured visual representations, such as sets of visual marks \citep{setofmark}, to encode interface components \citep{gpt4vinwonderland, androidinthezoo, wang2024gui}. Subsequent work has shifted toward end-to-end visual agents that directly predict actions from raw screenshots \citep{cocoagent, uground, autoglm}.

With the scaling of training data, recent agents have demonstrated strong capabilities in grounding natural language instructions to interface elements \citep{osatlas, ariaui, uitars}. This performance has been further enhanced by incorporating reinforcement learning techniques, including static RL methods\citep{uir1, segui, guig2, ui-venus} and online RL methods\citep{mai-ui, uitars2}. However, most existing approaches assume that the agent’s observation consists of a screenshot from a single device at any given time \citep{nguyen2025gui}, leaving the more realistic and challenging setting of cross-device GUI interaction largely unaddressed, which remains a critical barrier to real-world deployment.

\subsection{GUI Agent Benchmark}

Alongside the evolution of GUI agents, benchmarking methodologies have progressed from static, single-step grounding tasks \citep{seeclick, screenspotpro, transbench} to multi-step interactive settings \citep{aitw, guiodyssey}, and further to virtual environments supporting diverse execution paths \citep{osworld, crab, omnibench}. However, most benchmarks remain confined to single-device settings. Although CRAB \citep{crab} acknowledges the importance of cross-device task execution, its coverage remains limited: it contains only 18 cross-device tasks (largely due to annotation difficulty) and does not model realistic file and data transfer across devices, making it insufficient for evaluating end-to-end cross-device workflows. Consequently, the lack of large-scale, systematic cross-device benchmarks leaves a gap between lab evaluation and real-world deployment, motivating us to provide practical guidance for bringing GUI agents into real workflows.

%% file: sections/3_environment.tex

\section{Environment}
\label{environment}
A robust benchmark for cross-device GUI agents \textbf{necessitates} an evaluation environment that is reproducible, controllable, and capable of supporting both \textbf{high-throughput parallel execution} and \textbf{multi-device collaboration}. 
To fulfill these requirements, we propose the environment design of \jarvisbench. We will introduce the task definition and evaluation, the system architecture, and the environment configuration.

\subsection{Task Definition and Evaluation}
\label{task_definition}

Establishing a rigorous formal task definition is a prerequisite for constructing an environment that supports systematic evaluation. In \jarvisbench, a GUI agent task is modeled as a typed, parameterized system that supports dynamic instantiation, is executed by an agent in heterogeneous environments, and is evaluated through inspection of the final environment state.

A task is formally defined as a tuple $\mathcal{T} = (\mathcal{I}, \mathcal{O}, D, \Phi, P)$, consisting of a sequence of input slots $\mathcal{I}$ and output slots $\mathcal{O}$, a natural language template $D$, a set of evaluation functions $\Phi$, and a target execution platform $P$.
To support generalization, we distinguish between static specifications ($\mathcal{I}, P$) and dynamic components ($\mathcal{O}, D, \Phi$). The latter are materialized via a parameterized mapping $M$ conditioned on concrete input values $v_{\mathcal{I}}$ during execution, denoted as $(\mathcal{O}, D, \Phi) \leftarrow M(v_{\mathcal{I}})$.

\paragraph{Task Execution as State Transition.}
The execution of a GUI agent task is modeled as an iterative state transition process. At step $i$, the agent maintains an internal model state $M_{i-1}$ that encodes its memory from previous steps. Given the current environment observation $s_i$ and the task instruction $D$, the agent produces an action $a_i$ and updates its internal state:
\[
(a_i, M_i) = \mathrm{Agent}(s_i, D, M_{i-1}).
\]
The action $a_i$ is then executed in the environment, inducing a transition to the next environment state:
\[
s_{i+1} = \mathrm{Env}(s_i, a_i).
\]

In heterogeneous settings, the environment state $s_i$ may consist of observations from multiple platforms, i.e.,
\[
s_i = \{ s_i^{(P_1)}, s_i^{(P_2)}, \ldots, s_i^{(P_n)} \},
\]
allowing the agent to reason jointly over multiple environments within a single task execution. When the process terminates, the task reaches a final state $s_{\mathrm{final}}$, which is then evaluated by the corresponding evaluators in $\Phi$.

\paragraph{Dynamic Evaluation.}
An evaluator $\phi \in \Phi$ is a deterministic function $\phi: \mathcal{S} \rightarrow \{0,1\}$, where $\mathcal{S}$ denotes the environment state space. Given the final state $s_{\text{final}} \in \mathcal{S}$ after task execution, $\phi(s_{\text{final}})$ verifies task completion by programmatically inspecting observable artifacts (e.g., file content or DOM elements), ensuring robustness to stochastic interface changes.

\subsection{System Architecture}
\begin{figure}[t]
    \centering
    \includegraphics[width=\linewidth]{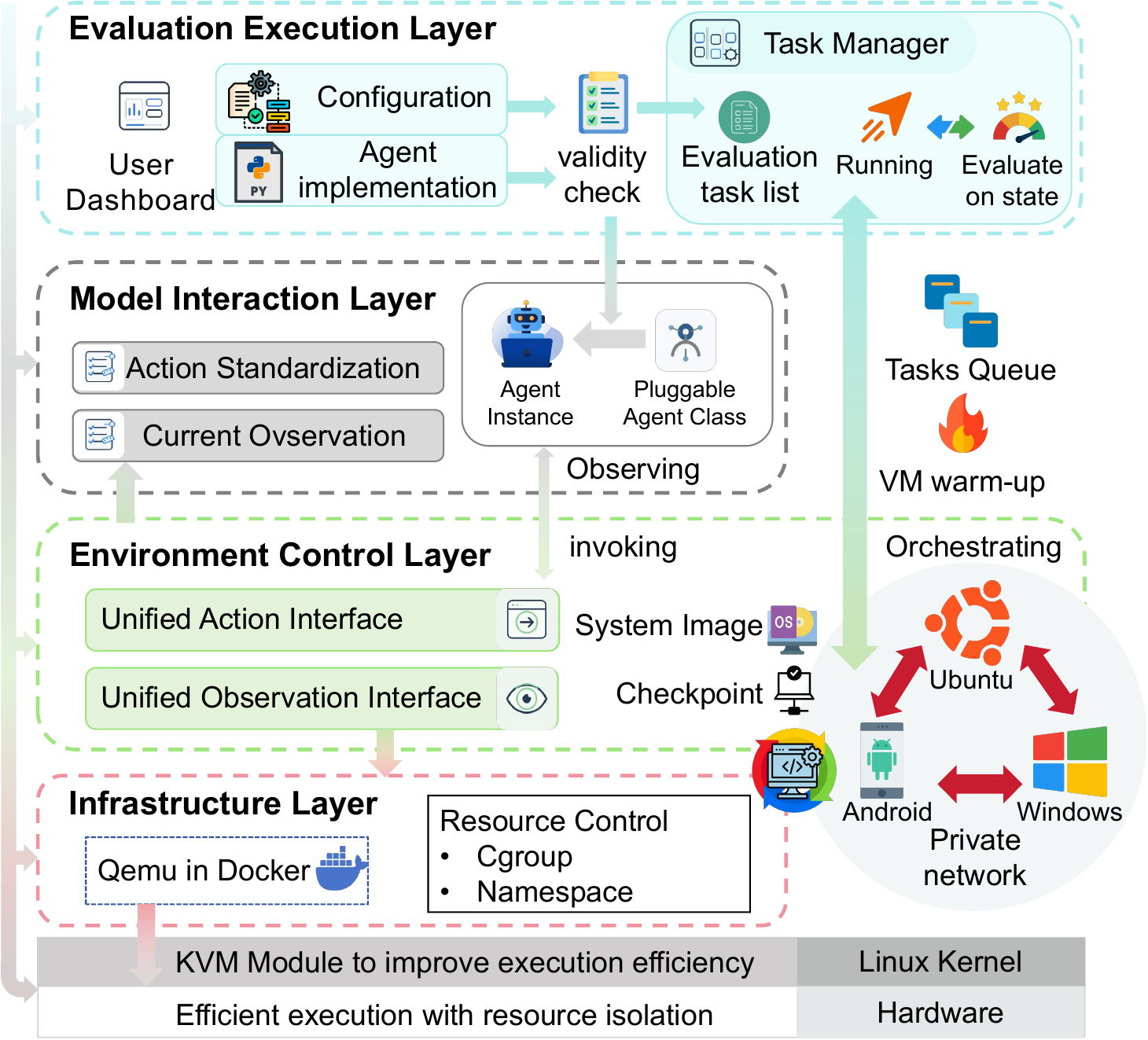} 
    \caption{The system architecture of \jarvisbench environment.}
    \label{fig:Architecture}
\end{figure}

The \jarvisbench environment is organized into four layers: the Infrastructure Layer, the Environment Control Layer, the Model Interaction Layer, and the Evaluation Execution Layer, as shown in Figure \ref{fig:Architecture}. Based on this layered design, \jarvisbench serves as a reproducible and controllable evaluation environment that supports high-throughput parallel execution and multi-device collaboration.

\paragraph{Infrastructure Layer}
The Infrastructure Layer aims to establish a reproducible experimental environment via Docker-based virtualization, which enables isolated execution with minimal overhead. It is responsible for handling the creation and destruction of Docker containers, exposing external interfaces for the lifecycle management of the Docker environment. 

Please refer to Appendix \ref{Infrastructure Details} for infrastructure details.

\paragraph{Environment Control Layer}
The Environment Control Layer is designed as a platform-agnostic abstraction layer. By defining a unified interface and action space, it masks the heterogeneity of underlying platforms, enabling standardized control over diverse device environments. Please refer to Appendix \ref{Environment Control Layer Details} for details of the unified observation space and the unified action space.

\paragraph{Model Interaction Layer}
The Model Interaction Layer is responsible for translating abstract natural language instructions into cross-platform executable action sequences. To address the complex decision-making required for multi-device tasks, this layer adopts a two-stage \textbf{Planner-Grounder architecture}. Additionally, it is tasked with standardizing and mapping diverse model inference results into our pre-defined action space. Please refer to Appendix \ref{sec:implementation-details} for further details.

\paragraph{Evaluation Execution Layer}
The Evaluation Execution Layer is charged with the full-lifecycle management of evaluation tasks. Beyond orchestrating the Infrastructure Layer to enable high-throughput parallel evaluation, it allows for the real-time supervision of agent-environment interactions and performs automated scoring based on predefined evaluation protocols.

The standard evaluation workflow proceeds as follows: The layer first parses the Task Configuration File (detailed in Section~\ref{Environment Configuration}) and provisions the necessary Docker containers and computing resources. Subsequently, it invokes the Environment Control Layer to perform pre-execution initialization (e.g., uploading files), while extracting the natural language instructions and loading the evaluation rules from the configuration. The system then enters an interaction-execution loop: the agent acquires observations and plans actions, which are dispatched to the target device. This loop persists until the task concludes. Finally, essential state data is retrieved to the host machine for automated scoring based on the rules to determine the task's completion status.

Please refer to Appendix \ref{Dashboard Details} for further details.

\subsection{Environment Configuration}
\label{Environment Configuration}
Given that individual tasks often entail specific execution prerequisites (e.g., ensuring a specific file pre-exists in the environment) and distinct evaluation criteria, we utilize a structured configuration file to encapsulate all task-specific metadata. This design enables automated management, where the Evaluation Execution Layer parses the file to dynamically apply the corresponding environment configurations. For further details regarding the configuration files, please refer to Appendix \ref{Environment Configuration Details}.

%% file: sections/4_JarvisBench.tex
\section{\jarvisbench}

\begin{figure*}[!ht]
    \centering
    \includegraphics[width=1\linewidth]{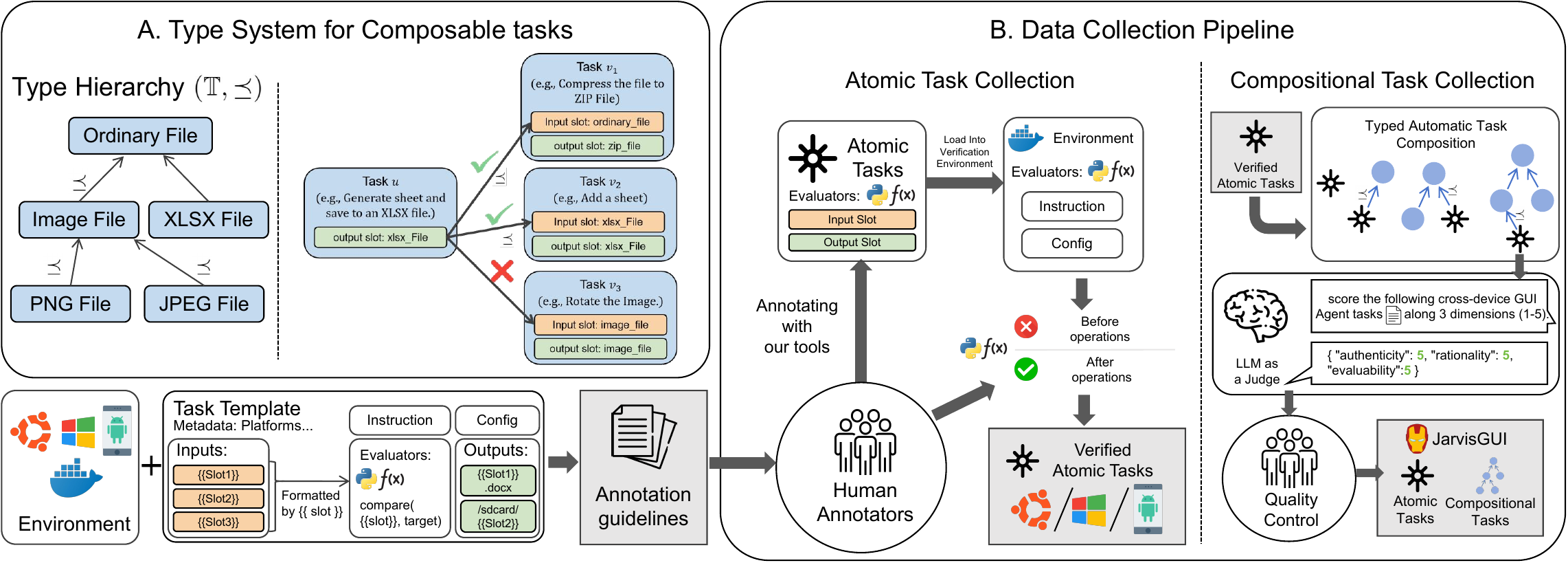}
    \caption{Data Collection Pipeline of \jarvisbench.}
    \label{fig:jarvisbench}
\end{figure*}

Built upon the task definition, evaluation protocol, and heterogeneous environment introduced in Section~\ref{environment}, we construct a task-generation pipeline for building \jarvisbench and comprehensively evaluating GUI agents in cross-device settings, as shown in Figure~\ref{fig:jarvisbench}.

We first introduce our type system and its role in compositional task construction (Section~\ref{compositional_task}). Next, we describe the data collection pipeline used to generate diverse and complex cross-device GUI tasks (Section~\ref{data_collection_pipeline}). Finally, we report our quality control methods (Section~\ref{quality_control}) and comprehensive statistics (Section~\ref{data_statistics}) of \jarvisbench.

Table \ref{tab:related_work} shows the detailed comparison between \jarvisbench and other popular benchmarks.

\input{tables/related_works}

\subsection{Type System and Compositional Task Modeling}
\label{compositional_task}
\paragraph{Type System} 
To enable automated validity checking for task composition, we introduce a hierarchical type system.
Let $\mathbb{T}$ be a type space equipped with a partial order $\preceq$ representing the subtype relation. Each slot $s \in \mathcal{I} \cup \mathcal{O}$ is defined as a tuple $(\mathrm{id}, \tau)$, where $\tau \in \mathbb{T}$.
Data flow validity is enforced via subtype compatibility: an output slot with type $\tau_{\mathrm{out}}$ may satisfy an input slot with type $\tau_{\mathrm{in}}$ if and only if $\tau_{\mathrm{out}} \preceq \tau_{\mathrm{in}}$. This typing discipline ensures semantic consistency when chaining specialized outputs (e.g., \texttt{xlsx\_file}) to more general inputs (e.g., \texttt{ordinary\_file}).

\paragraph{Task Graph}
We model a complex GUI task workflow as a structured composition of task instances, organized as a directed graph $\mathcal{G} = (\mathcal{V}, \mathcal{E})$. Each node $v \in \mathcal{V}$ corresponds to a concrete instantiation of a task
\[
\mathcal{T}_v = (\mathcal{I}_v, \mathcal{O}_v, D_v, \Phi_v, P_v),
\]
where the dynamic components $(\mathcal{O}_v, D_v, \Phi_v)$ are instantiated via the mapping $(\mathcal{O}_v, D_v, \Phi_v) \leftarrow M(v_{\mathcal{I}_v})$ at execution time.  
A directed edge $(u, v) \in \mathcal{E}$ indicates that the outputs $\mathcal{O}_u$ of task $u$ are provided as inputs $\mathcal{I}_v$ to task $v$, subject to type compatibility.

\paragraph{Global Success Criterion}
The execution of a workflow is deemed successful if and only if, for every task instance $v \in \mathcal{V}$, all its associated evaluators are satisfied on the final environment state:
\begin{equation}
\label{eq:global_success}
\mathrm{Success}(\mathcal{G}) \iff 
\forall v \in \mathcal{V}, \; \forall \phi \in \Phi_v, \; \phi(s^v_{\mathrm{final}}) = 1 .
\end{equation}

\subsection{Data Collection Pipeline}
\label{data_collection_pipeline}

We construct \jarvisbench through a multi-stage data collection and task composition pipeline that combines human annotation, automated task synthesis, and quality filtering.

\paragraph{Template Task Annotation}
Following prior work\citep{osworld}, we first employ experienced human annotators to design dozens of template tasks for each target platform. Each annotated task is required to be input-agnostic, such that it can accept arbitrary user-provided inputs. This property is achieved by introducing textual placeholders (e.g., \{\{input\_slot\_name\}\}) into the dynamic components of task descriptions, which enables flexible instantiation with different input values at runtime.

Special care is taken in the design of task evaluators. We require all evaluators to be implemented as parameterized Python functions, where evaluation criteria are likewise specified using textual placeholders. This design allows evaluators to be dynamically formatted according to instantiated input slots, ensuring that evaluation logic remains consistent and robust across diverse task instances.

\paragraph{Sampling-Based Automatic Task Composition}
To construct compositional tasks at scale, we adopt a sampling-based procedure that automatically stitches individual task instances into a directed task graph, as illustrated in Algorithm~\ref{alg:task_stitching}. By leveraging our formally defined type system and dynamic task instantiation mechanism, the composed task graphs are guaranteed to be well-typed and valid in the majority of cases.

For task compositions involving cross-device file transfer, we additionally introduce auxiliary file-transfer tasks that move files to temporary directories on the target device.

\paragraph{Value Propagation}
For a composed task graph $\mathcal{G} = (\mathcal{V}, \mathcal{E})$, values produced by upstream tasks are propagated along directed edges to downstream tasks. Specifically, for an edge $(u, v) \in \mathcal{E}$ that connects an output slot of $u$ to a compatible input slot of $v$, the runtime value is propagated and used to instantiate the dynamic components of task $v$ via the parameterized mapping $M$, including task descriptions and evaluation functions. This mechanism ensures that downstream tasks are conditioned on concrete upstream results rather than static placeholders.

After obtaining the natural language descriptions of all subtasks, we rewrite the merged instructions using Qwen3-32B, instructing the model to strictly preserve all procedural details while generating a concise, coherent, and unambiguous final task description.

\paragraph{LLM-as-a-Judge Evaluation and Filtering}
While our type system guarantees that most automatically generated tasks are well-typed and structurally valid, formal correctness alone does not ensure alignment with real user needs. Following prior work \citep{xia2025safetoolbench,cheng2025toolspectrum,wang2024appbench}, we therefore sample 2,000 tasks and evaluate them using an LLM-as-a-judge framework along three dimensions:
(1) \textit{Task realism}: whether the task reflects a plausible real-world user need;
(2) \textit{Task coherence}: whether the task design is logically consistent and free of superfluous or meaningless steps;
(3) \textit{Evaluability}: whether the task has well-defined objectives and produces outputs that can be reliably assessed.

Each dimension is rated on a 5-point scale. Only tasks receiving the maximum score (5) on all three criteria are retained, forming the final high-quality task set in \jarvisbench.

The judge model was Qwen3-30B-A3B-Thinking-2507-FP8, which provides a practical balance between evaluation accuracy and computational cost. 

\subsection{Quality Control}
\label{quality_control}
We adopt a multi-stage quality control process to ensure the correctness and reliability of \jarvisbench. For human-annotated atomic tasks, we enforce strict guidelines: each task must be completable by a human annotator within 20 interaction steps in our virtual environment based solely on the task instruction. Task evaluators are automatically executed both before and after task completion, and only tasks for which the evaluator reports incomplete before execution and complete after execution are retained. For all retained atomic tasks, we further perform manual static inspection to verify the correct usage of all textual placeholders in task definitions. 

After automated task composition, following prior work \citep{liu2025repodebug, deng2025retail,liu2021durecdial}, we randomly sampled and manually inspected 50 composed tasks. To ensure rigorous human validation, this expert evaluation assessed the tasks along the same three dimensions used in the LLM filtering (task realism, task coherence, and evaluability). Each task was first reviewed by one author, while unclear or preference-sensitive cases were jointly discussed by all authors until a consensus decision was reached. We found that 92\% of the tasks were fully correct and reflected concrete real-world needs, while 4 tasks had imprecise descriptions that might introduce ambiguity. To further clarify the reliability of the LLM-based filtering process, we calculated the agreement between the LLM filtering decisions and human judgments using Cohen’s Kappa. The resulting Cohen’s Kappa score is approximately 0.92, indicating strong agreement between the LLM filter and human evaluation.

Furthermore, our preliminary analysis of tasks rejected by the LLM filter shows several recurring groups of common composition failures: unevaluable workflows, semantically incoherent compositions, and tasks with weak realism.

\subsection{Data Statistics}
\label{data_statistics}
\input{tables/data_statistics}

Table~\ref{tab:data_statistic} summarizes the statistics of our benchmark, which consists of both single-platform atomic tasks and compositional tasks. By applying our data collection pipeline, we obtain 150 compositional tasks, evenly distributed across three categories: single-platform tasks with dependencies, cross-platform tasks without dependencies, and cross-platform tasks with dependencies (50 tasks each).

Each compositional task can be further decomposed into platform-specific subtasks, resulting in a total of 442 subtasks. Among them, 187 subtasks are executed on Ubuntu, 138 on Windows, and 52 on Android, while the remaining 65 subtasks correspond to file transfer assistant tasks that facilitate cross-platform coordination. More detailed statistics can be found in Appendix~\ref{sec:task_detailed_statistics}.

%% file: tables/related_works.tex
\newcommand{\partialyes}{%
\textcolor{orange}{%
\ding{51}\kern-0.75em\raisebox{0.1em}{\scriptsize$\bullet$}%
}%
}
\begin{table}[t]

\centering
\tabcolsep=0.15cm
\begin{adjustbox}{max width=0.48 \textwidth}
\begin{tabular}{l|c c c c c}
\hline
\textbf{Benchmark}
& \textit{CD} 
& \textit{MS} 
& \textit{DE} 
& \textit{DT} 
& \textit{\#Devices} \\
\hline

screenspot \citep{seeclick} 
& \color{deepred}\XSolidBrush 
& \color{deepred}\XSolidBrush 
& \color{deepred}\XSolidBrush 
& \color{deepred}\XSolidBrush 
& 4
\\

screenspot-pro \citep{screenspotpro} 
& \color{deepred}\XSolidBrush 
& \color{deepred}\XSolidBrush 
& \color{deepred}\XSolidBrush 
& \color{deepred}\XSolidBrush 
& 3
\\

TransBench \citep{transbench} 
& \color{deepred}\XSolidBrush 
& \color{deepred}\XSolidBrush 
& \color{deepred}\XSolidBrush 
& \color{deepred}\XSolidBrush 
& 3
\\

AndroidControl \citep{androidcontrol} 
& \color{deepred}\XSolidBrush 
& \color{deepgreen}\CheckmarkBold 
& \color{deepred}\XSolidBrush 
& \color{deepred}\XSolidBrush 
& 1
\\

AITW \citep{aitw} 
& \color{deepred}\XSolidBrush 
& \color{deepgreen}\CheckmarkBold 
& \color{deepred}\XSolidBrush 
& \color{deepred}\XSolidBrush 
& 1
\\

GUI Odyssey \citep{guiodyssey} 
& \color{deepred}\XSolidBrush 
& \color{deepgreen}\CheckmarkBold 
& \color{deepred}\XSolidBrush 
& \color{deepred}\XSolidBrush 
& 5
\\

OS World \citep{osworld} 
& \color{deepred}\XSolidBrush 
& \color{deepgreen}\CheckmarkBold 
& \color{deepgreen}\CheckmarkBold 
& \color{deepred}\XSolidBrush 
& 3
\\

CRAB \citep{crab} 
& \raisebox{0.8ex}{\color{orange}$\sqrt{}\mkern-9mu{\smallsetminus}$}
& \color{deepgreen}\CheckmarkBold 
& \color{deepgreen}\CheckmarkBold 
& \color{deepred}\XSolidBrush 
& 2
\\

OmniBench\citep{omnibench} 
& \color{deepred}\XSolidBrush 
& \color{deepgreen}\CheckmarkBold 
& \color{deepgreen}\CheckmarkBold 
& \color{deepred}\XSolidBrush 
& 3
\\

\hline
\textbf{\textsc{\jarvisbench}} (Ours) 
& \color{deepgreen}\CheckmarkBold 
& \color{deepgreen}\CheckmarkBold 
& \color{deepgreen}\CheckmarkBold 
& \color{deepgreen}\CheckmarkBold 
& \textbf{3} 
\\
\hline
\end{tabular}
\end{adjustbox}

\caption{
Comparison of \jarvisbench with existing GUI agent benchmarks. CD: cross-device task execution; DE: evaluation based on final states of a dynamic environment; DT: dynamically generated task instructions and evaluation criteria; MS: single task requires multiple steps.
}
\label{tab:related_work}


\end{table}

%% file: tables/data_statistics.tex
\begin{table}[t]
\centering

    \resizebox{0.85\linewidth}{!}{
\begin{tabular}{l c c}
\toprule
\textbf{Category} & \textbf{\#Task} & \textbf{\#APP} \\
\midrule
{\textit{Single-platform atomic tasks}} & 118 & 31\\
\quad Android & 24 & 10 \\
\quad Ubuntu & 56 & 11\\
\quad Windows & 38 & 10\\
\midrule
\textit{Compositional tasks} & 150 & 29 \\
\quad Single-platform with dependencies & 50 & 12\\
\quad Cross-platform without dependencies & 50 & 26\\
\quad Cross-platform with dependencies & 50 & 18\\
\midrule
\quad subtasks of Compositional tasks & 442 & 29\\
\quad\quad Ubuntu & 187 & 11\\
\quad\quad Windows & 138 & 8\\
\quad\quad Android & 52 & 10\\
\quad\quad File transfer assistant tasks & 65 & / \\
\bottomrule
\end{tabular}
}

\caption{Data statistics of \jarvisbench}
\label{tab:data_statistic}
\vspace{-10pt}
\end{table}

%% file: sections/5_experiment.tex
\section{Experiments}
\input{tables/main_results}

\subsection{Experiment Setting}
\paragraph{Baselines}
Following prior work~\citep{omnibench}, we conduct comprehensive experiments on \textsc{\jarvisbench} using Qwen3-VL-30B-A3B-Instruct~\citep{qwen3vl}, HOLO2-30B-A3B~\citep{holo2}, UI-TARS-1.5-7B~\citep{uitars}, MAI-UI-8B~\citep{mai-ui}, UI-Venus-Ground-7B~\citep{ui-venus}, and GUI-Owl-32B~\citep{gui-owl}.

\paragraph{Metrics}
To rigorously evaluate GUI-agent performance across different task 
granularities and platforms, we organize our experimental setting into two primary categories encompassing six specific scenarios: \textbf{(1) Atomic Tasks}, utilized to assess variable-parameter execution robustness on \texttt{Android}, \texttt{Windows}, and \texttt{Ubuntu}. \textbf{(2) Multi-Tasks}, aimed at evaluating complex planning capabilities through single-device dependent (\textbf{SW}), multi-device independent (\textbf{MI}), and multi-device dependent (\textbf{MD}) workflows. We employ a tailored evaluation protocol where metrics are applied based on task granularity.



\begin{itemize}[leftmargin=*]
    \item \textbf{Total Task Success Rate (TSR)} serves as the universal metric for \textbf{both Atomic and Multi Tasks}. It evaluates the agent's holistic effectiveness by calculating the proportion of tasks where the terminal state strictly aligns with the user's goal. Formally, TSR is defined as:
    \begin{equation}
        \text{TSR} = \frac{1}{|\mathcal{T}|} \sum_{t \in \mathcal{T}} \mathbbm{1}_{\text{success}(t)},
    \end{equation}
    where $\mathcal{T}$ is the evaluation task set and $\mathbbm{1}_{\text{success}(t)}$ equals $1$ if task $t$ is successfully completed, and $0$ otherwise.

    \item \textbf{Sub-task Success Rate (SSR)} is introduced \textbf{specifically for Multi-Tasks} to capture the agent's reliability in long-horizon planning. Unlike the binary outcome of TSR, SSR provides fine-grained insights by tracking the average completion ratio of intermediate steps within complex workflows. We formulate SSR as:
    \begin{equation}
        \text{SSR} = \frac{1}{|\mathcal{T}_{\text{comp}}|} \sum_{t \in \mathcal{T}_{\text{comp}}} \frac{|s_{t}^{\text{passed}}|}{|s_{t}^{\text{total}}|},
    \end{equation}
    where $\mathcal{T}_{\text{comp}}$ denotes the compositional task set, and $|s_{t}^{\text{passed}}|$ and $|s_{t}^{\text{total}}|$ denote the numbers of completed and required sub-tasks, respectively.
\end{itemize}

\paragraph{Implementation Details}
To effectively align and process multi-source visual inputs from heterogeneous devices, we adopt a hierarchical \textbf{Planner--Grounder architecture} as a representative instantiation in our study. Existing specialized GUI models are typically post-trained on single-device GUI agent datasets, which limits their generalization capability and prevents them from jointly handling high-level action planning across devices and low-level action grounding (e.g., coordinate prediction). To overcome these limitations, we employ the general-purpose VL model Qwen3-VL-Plus~\citep{qwen3vl} as a centralized planner that performs high-level reasoning and cross-platform coordination.

The planner receives real-time screenshots from three platforms (Android, Windows, and Ubuntu), together with the user instruction, while maintaining the interaction history as contextual memory. Guided by the prompts shown in Table~\ref{tab:planner_prompt}, the planner follows a structured \textit{Observation--Planning--Action} reasoning protocol to determine the next abstract action and the target operating platform. For atomic actions that require precise GUI element localization, such as \texttt{clicking} or \texttt{scrolling}, the planner delegates execution to a specific \textbf{Grounding Agent}, which predicts the exact coordinates conditioned on the planner's high-level intent.

\subsection{Main Results}
Table \ref{tab:main_results} presents the comparative results of different models, we draw the following conclusions.

\paragraph{\textit{Cross-Device Dependencies Create Fragile Critical Paths.}}
The most significant performance drop is observed in multi-device dependent \textbf{(MD)} tasks, where the agents struggle to achieve a non-zero success rate, lagging significantly behind independent tasks. We attribute this failure to two primary factors. First, the \textbf{state inference bottleneck}: the agent finds it difficult to correctly infer the current environmental state of a target device (e.g., the precise location of a file) solely based on historical execution records from another device. Second, the \textbf{long-horizon critical path}: these tasks often involve strictly sequential steps (such as file transmission followed by processing). A single failure in an intermediate step—such as an inability to locate a file or a login timeout during cloud transfer—breaks the entire chain, rendering the subsequent planning futile.

\paragraph{\textit{Multi-Device Contexts Impose Cognitive and Visual Overload.}}
Even in scenarios where sub-tasks are independent (MI), the coordination of multiple devices leads to a notable decline in success rates compared to single-device settings. This degradation stems from the complexity of the observation space; processing three simultaneous screens reduces the model's visual grounding accuracy. Furthermore, the model exhibits limitations in \textbf{contextual reasoning} within complex environments. It often fails to map implicit instructions to the correct platform—for instance, inferring that "C drive" implies Windows or that "back to the computer" refers to the previously operated system—leading to incorrect platform routing despite the independence of the tasks.

\paragraph{\textit{Logical Dependencies Hinder Workflow Completion.}}
Performance on single-device dependent tasks (SW) is consistently lower than that on independent or atomic tasks, highlighting the gap between action execution and workflow planning. This indicates that even when the agent possesses the capability to execute atomic actions with variable parameters (dynamic slots), correctly managing \textbf{inter-task dependencies} remains a hurdle. The difficulty lies in preserving intermediate results (e.g., maintaining clipboard content or temporary file paths) and sequencing actions logically, which is significantly more demanding than isolated command execution.

\paragraph{\textit{Platform Bias Exists in Atomic Action Robustness.}}
While atomic tasks present a general challenge across the board, there is a distinct performance disparity among operating systems. The evaluated models consistently perform better on \textbf{Ubuntu} and \textbf{Android} compared to \textbf{Windows}. We attribute this phenomenon to the bias in pre-training data distributions. The abundance of mobile interaction data (for Android) and command-line/script-heavy web data (for Ubuntu) in general corpora likely provides stronger supervision, whereas high-quality, diverse GUI interaction data for Windows is relatively scarcer, resulting in weaker generalization on Windows-specific controls.

To provide rigorous statistical evaluations and verify that the observed cross-device bottlenecks generalize across different planner models, detailed confidence intervals and supplementary experiments on additional mainstream models are provided in Appendix \ref{sec:experiment_details}.

\subsection{Error Analysis}

We identify several recurring system-level failure modes that appear consistently across both atomic and compositional tasks. These errors typically arise from incomplete execution accompanied by incorrect assumptions about internal state. Common manifestations include premature termination before all instruction requirements are met, omission of essential intermediate or follow-up actions, and incorrect selection or sequencing of low-level operations. 

\begin{figure}[!ht]
    \centering
    \includegraphics[width=1\linewidth]{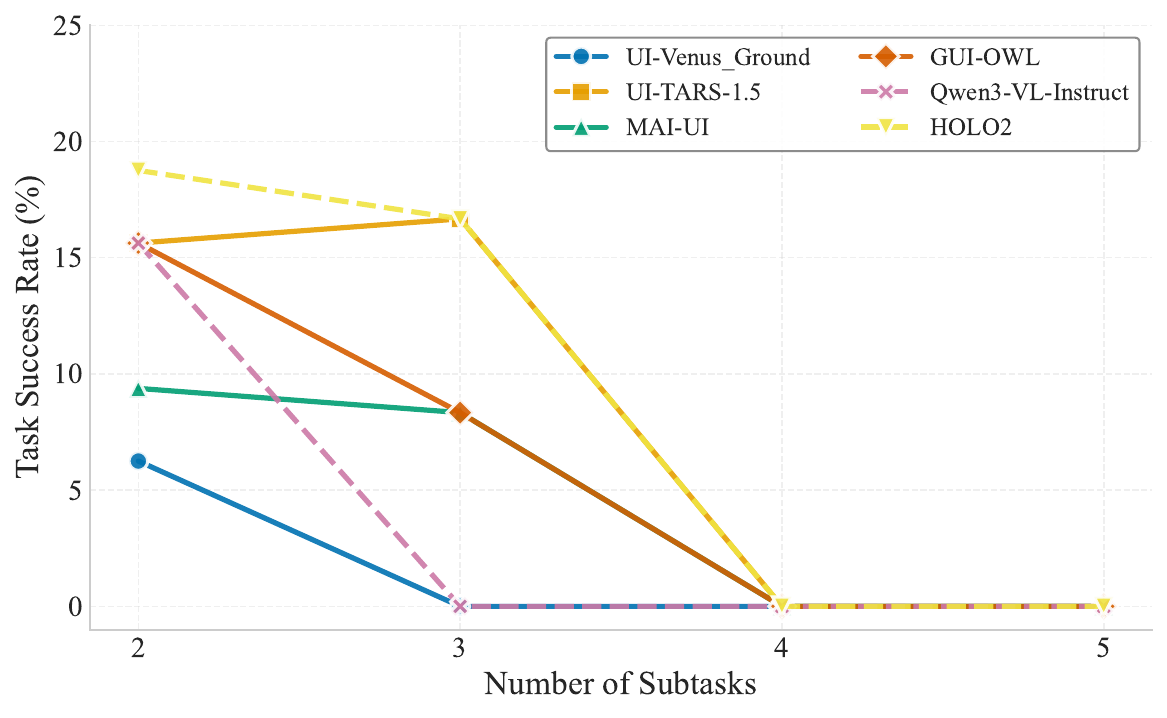}
    \caption{Task success rate vs. number of subtasks for composite single-device tasks.}
    \label{fig:task_success_tread_icml}
\end{figure}

Crucially, these isolated errors compound rapidly as task complexity grows. As illustrated in Figure~\ref{fig:task_success_tread_icml} and Appendix~\ref{sec:failure_mode}, the task success rate exhibits a sharp decline as the number of subtasks increases, dropping to near zero for tasks requiring four or more subtasks across all evaluated models. This degradation highlights a critical limitation in long-horizon planning: dependency steps that were never executed (e.g., file transfer or renaming) are nonetheless treated as completed, producing cascading errors in subsequent actions. Detailed successful and failed case studies are provided in Appendix~\ref{appendix_case_study}.

%% file: tables/main_results.tex
\newcommand{\best}[1]{\textbf{#1}}

\begin{table*}[!ht]
    \centering
    \small

    \renewcommand\arraystretch{1.2} 
    
    \resizebox{1\linewidth}{!}{
        \begin{tabular}{l c cc cc cc cc cc cc}
            \toprule
            
            \multirow{3}{*}{\textbf{Model}} & \multirow{3}{*}{\textbf{Size}} 
            & \multicolumn{4}{c}{\textbf{Atomic Tasks}} & \multicolumn{8}{c}{\textbf{Multi-Tasks}} \\
            
            \cmidrule(lr){3-6} \cmidrule(lr){7-14}
            
             & 
             & \multicolumn{1}{c}{Android} & \multicolumn{1}{c}{Windows} & \multicolumn{1}{c}{Ubuntu} &  \multicolumn{1}{c}{Overall}
             & \multicolumn{2}{c}{SW} & \multicolumn{2}{c}{MI} & \multicolumn{2}{c}{MD} & \multicolumn{2}{c}{Overall} \\
            
            \cmidrule(lr){7-8} 
            \cmidrule(lr){9-10} \cmidrule(lr){11-12} \cmidrule(lr){13-14}
            
             &  
             & TSR & TSR & TSR & TSR & TSR & SSR 
             & TSR & SSR & TSR & SSR & TSR & SSR \\
            \midrule

UI-Venus\_Ground\citep{ui-venus}    & 7B        
& 33.3 & \best{18.4} & 51.8 & 37.3 
& 4.0 & 23.2 & 0.0 & 24.0 & 0.0 & 9.7 & 1.3 & 16.7 \\

UI-TARS-1.5\citep{uitars}         & 7B        
& \best{50.0} & \best{18.4} & 44.6 & 37.3
& 14.0 & \best{28.8} & 4.0 & 28.0 & 0.0 & 8.8 & 6.0 & 18.8 \\

MAI-UI\citep{mai-ui}              & 8B        
& 37.5 & 7.9 & 39.3 & 28.8 
& 8.0 & 24.0 & 4.0 & 27.0 & 0.0 & 6.5 & 4.0 & 16.1 \\

GUI-OWL\citep{gui-owl}             & 32B       
& 41.7 & \best{18.4} & 51.8 & 39.0 & 12.0 & \best{28.8} & \best{8.0} & \best{29.0} & \best{2.0} & 6.9 & 7.3 & 18.1 \\

Qwen3-VL-Instruct\citep{qwen3vl}   & 30B (A3B) 
& 37.5 & 13.2 & 50.0 & 35.6 
& 10.0 & 24.0 & 2.0 & 28.0 & 0.0 & 10.1 & 4.0 & 18.1 \\

HOLO2\citep{holo2}               & 30B (A3B) 
& 41.7 & 15.8 & \best{60.7} & \best{42.4} 
& \best{16.0} & 28.0 & 6.0 & 26.0 & \best{2.0} & \best{10.6} & \best{8.0} & \best{19.0} \\

            \bottomrule
        \end{tabular}
    }

        \caption{
    \textbf{Main Experimental Results.} 
    Performance comparison on Atomic and Multi-task benchmarks. 
    \textbf{TSR}: Total Task Success Rate; \textbf{SSR}: Sub-task Success Rate.
    The best results are highlighted in \textbf{bold}.
    }
    \label{tab:main_results}
\end{table*}

%% file: sections/6_Conclusion.tex
\section{Conclusion}
We introduce \jarvisbench, a dynamic benchmark for evaluating GUI agents on diverse cross-device workflows across Android, Windows, and Ubuntu. By composing atomic tasks and evaluating final environment states, \jarvisbench reveals that current agents remain weak in state transfer, cross-platform reasoning, and long-horizon dependency management, highlighting key challenges for building reliable real-world GUI agents.

%% file: appendix.tex
\section{Implementation Details}
\label{sec:implementation-details}

\subsection{Planner Model}
\label{Planner-Model-Details}
To handle multi-device visual inputs, we introduce a two-stage planner–grounder framework and design task-consistent prompts tailored for cross-device scenarios, as shown in Table~\ref{tab:planner_prompt}. We adopt a two-stage architecture because most current GUI agents with publicly available inference implementations assume interaction with a single active platform at each step, rather than simultaneously receiving observations from multiple devices. This convention also applies to recent end-to-end agents such as UI-TARS-2 and AutoGLM: their end-to-end interaction loops are still defined within a single active GUI environment, as reflected in their published formulations and official inference implementations. Planner–grounder architecture decouples each interaction into two parts: (1) platform identification and high-level action selection from screenshots of multiple devices, and (2) action prediction from the screenshot of the selected platform. During execution, \{\{USER QUERY\}\} is formatted into task instructions at execution time, while \{\{CURRENT\_HISTORY\}\} contains the step\_plan from the model’s preceding steps. We additionally incorporate execution errors generated by the framework (e.g., “action not found”) into the current history, enabling the planner model to receive feedback from the framework when it produces invalid actions.

\begin{prompttable}{tab:planner_prompt}
{Prompt Template for Cross-Device GUI Planner}{Cross-Device GUI Planner Prompt}
This prompt defines a minimal, strict planner for a cross-device GUI Agent system.

Supported platforms: Android, Windows, Ubuntu. 

The planner's primary constraint is: 

When an action requires coordinates, the planner must NEVER output coordinates.

Coordinates are provided later by a GUI Grounding sub-agent. 
\\
\promptsection{1. Planner Role (STRICT)} 

You are a GUI task planner, not an executor. 

At each step, you receive three screenshots: Android, Windows, Ubuntu. 

Your job is to output exactly one next step. 
\\
\promptsection{2. What You Must Decide}

For every step: 

1. Read the screenshots and understand what is shown in each platform. 

2. Which single platform acts next.

3. What low-level action should be performed on the selected platform. 

4. Which tool action to invoke.

You must NOT: Infer or guess coordinates, describe pixel locations, or output more than one step. 
\\
\promptsection{3. Output Format (MANDATORY)} 

Output only one JSON object in \texttt{<step> ... </step>}: 

\texttt{<step>} 

\texttt{\{} 

\hspace*{1em} \texttt{"current\_status": "What is shown in the screenshots...",} 

\hspace*{1em} \texttt{"platform": "android | windows | ubuntu",} 

\hspace*{1em} \texttt{"step\_plan": "A short plan for this step to achieve the goal.",} 

\hspace*{1em} \texttt{"element\_description": "Instruction of the target element of this step, } 

\hspace{1em} \texttt{will passed to the GUI Grounding sub-agent to generate coordinates if needed.",}

\hspace*{1em} \texttt{"tool": \{} 

\hspace*{2em} \texttt{"action": "..."} 

\hspace*{1em} \texttt{\}} 

\texttt{\}} 

\texttt{</step>} 

\texttt{step\_plan} must: 

Describe the plan for this step in a short sentence; \\
Be clear and concise. \\
Good examples: \\
- "I need to first try to open Terminal", \\
- "I have to close this window". 

\texttt{element\_description} must: \\
Describe the target element of this step in a short sentence; \\
Be short and imperative. \\
Good examples: \\
- "Click on the Settings application icon", \\
- "Option list to scroll down". 
\\
\promptsection{4. Waiting and Termination} 
\\
If UI is loading or changing $\rightarrow$ use \texttt{wait}. \\
When task finishes: \texttt{\{ "action": "terminate", "status": "success" \}} 
\\
\promptsection{5. History Summary (READ-ONLY)} \\
At each step, you are given a history summary. 

\texttt{Step 1 on <platform>: <step1\_plan>}\\
\texttt{Step 2 on <platform>: <step2\_plan>}\\
\texttt{...}

Rules: History is read-only; Do not repeat completed steps; Do not output history; Step in the history may failed due to the sub-agent, you can try to execute in another way. \\
\promptsection{6. Tool Definitions} \\
This section explicitly lists which actions are allowed and which fields the planner is allowed to fill. \\
\\
\textbf{6.1 Android Allowed Actions} \\
The planner may choose one of the following actions: \\
-- \texttt{type}: Input the specified text into the activated input box. Fields: \texttt{action}, \texttt{text}. \\
-- \texttt{system\_button}: Press the system button (Examples: "home", "back", "menu", "enter", "volume\_up", "volume\_down"). Fields: \texttt{action}, \texttt{button}. \\
-- \texttt{wait}: Wait specified seconds for the change to happen. Fields: \texttt{action}, \texttt{time}. \\
-- \texttt{click}: Click on the specified element. \\
-- \texttt{long\_press}: Long press on the specified element. \\
-- \texttt{swipe}: Swipe from the specified element to four directions (Examples: "up", "down", "left", "right"). Fields: \texttt{action}, \texttt{direction}. \\
-- \texttt{terminate}: Stop the current task. Fields: \texttt{action}, \texttt{status}. \\
\\
Fields the Planner MAY Fill \\
\begin{tabular}{@{}ll@{}}
\texttt{type} & : \texttt{action}, \texttt{text} \\
\texttt{system\_button} & : \texttt{action}, \texttt{button} \\
\texttt{wait} & : \texttt{action}, \texttt{time} \\
\texttt{terminate} & : \texttt{action}, \texttt{status} \\
\texttt{swipe} & : \texttt{action}, \texttt{direction} \\
\end{tabular} \\
\\
Coordinate-Controlled Actions (STRICT): \\
For \texttt{click}, \texttt{long\_press}, \texttt{swipe}: \\
Planner MUST output only: \texttt{\{ "action": "click" \}} \\
Planner MUST NOT output \texttt{coordinate} or mention locations in \texttt{step\_instruction}.\\ Coordinates will be supplied by the GUI Grounding agent. \\
\\\textbf{6.2 Windows / Ubuntu Allowed Actions} \\
The planner may choose one of the following actions: \\
-- \texttt{type}: Input the specified text into the activated input box. Fields: \texttt{action}, \texttt{text}. \\
-- \texttt{key}: Performs key down presses on the arguments passed in order. Fields: \texttt{action}, \texttt{keys}. \\
-- \texttt{scroll}: Scroll the mouse wheel. Fields: \texttt{action}, \texttt{pixels}. \\
-- \texttt{hscroll}: Scroll the mouse wheel horizontally. Fields: \texttt{action}, \texttt{pixels}. \\
-- \texttt{wait}: Wait specified seconds for the change to happen. Fields: \texttt{action}, \texttt{time}. \\
-- \texttt{terminate}: 
\\

Fields the Planner MAY Fill (Summary) \\
\begin{tabular}{@{}ll@{}}
\texttt{type} & : \texttt{action}, \texttt{text} \\
\texttt{key} & : \texttt{action}, \texttt{keys} \\
\texttt{scroll/hscroll} & : \texttt{action}, \texttt{pixels} \\
\texttt{wait} & : \texttt{action}, \texttt{time} \\
\texttt{answer} & : \texttt{action}, \texttt{text} \\
\texttt{terminate} & : \texttt{action}, \texttt{status} \\
\end{tabular} \\

Coordinate-Controlled Actions (STRICT) \\
For the actions listed above (mouse actions): \\
-- Planner MUST output ONLY the action name, e.g., \texttt{\{ "action": "left\_click" \}}. \\
-- Planner MUST NOT output \texttt{coordinate}. \\
-- Planner MUST NOT mention positions in \texttt{element\_description}. 
\\
\promptsection{7. Transfer File Between Android and Windows / Ubuntu} \\
You have access to a alist server. \\
-- url: http://\{\{ALIST\_SERVER\_IP\}\}:5244 \\
-- username: agent \\
-- password: agentPassword \\
-- folder: /data \\
\\
You can upload files to the alist server or download files from the alist server by browser on all platforms. \\
-- first, open this url in browser: http://\{\{ALIST\_SERVER\_IP\}\}:5244 \\
-- then, login with username: agent and password: agentPassword \\
-- select the folder /data \\
-- select file to download, or click bottom-right "three dot button" to find the upload file button to upload file. 
\\
\promptsection{FINAL REMINDER} \\
-- Planner selects \textbf{platform + action + step\_plan + element\_description} only. \\
-- Executor + Grounder handle \textbf{coordinates and execution}. \\
-- If you have already performed two identical operations in history, please change the execution method. \\
\\
Current history: \\
\texttt{\{\{CURRENT\_HISTORY\}\}} \\
\\
User Query: \texttt{\{\{USER\_QUERY\}\}} \\

\end{prompttable}

\subsection{Data Filtering}

To select the highest-quality tasks from randomly generated candidates, we prompt a large language model to evaluate each task from three aspects and retain only tasks that receive full scores, as shown in Table~\ref{tab:rating_prompt}. This data filtering step is necessary because type compatibility only ensures structural validity, whereas LLM-based filtering further checks semantic coherence and evaluability. Our diagnostic analysis shows that removing the LLM filter would retain a substantial number of samples that do not satisfy the benchmark’s evaluation requirements. For example, one rejected task first deleted a screenshot and then required the same file to be copied and transferred, rendering the workflow unevaluable.

\begin{prompttable}{tab:rating_prompt}
{Prompt Template for LLM as a Judge to rating the generated tasks}{Prompt Template for LLM as a Judge}

Please score the following cross-device GUI Agent tasks. Each task will sequentially describe the input, commands to be executed, and the output. Inputs and outputs are represented by \texttt{\{\{element\}\}} slots. There may be dependencies between tasks, such as "from \texttt{\{\{another\_output\}\}} as \texttt{\{\{current\_input\}\}}". \\
\\
Please rate from the following three dimensions (1-5): \\
1. \textbf{Task Authenticity}: Whether it reflects real user needs and scenarios. \\
2. \textbf{Task Rationality}: Whether the task design is logical and avoids meaningless operations. For example, "compressing and copying a task output to another device" is rational, whereas "sending an Android output to Windows only to put it in the Recycle Bin" is irrational because the second step serves no purpose. Using random text as a replacement source for an \texttt{.xlsx} file is irrational (as a match is nearly impossible), but using it as a replacement target is rational. \\
3. \textbf{Evaluability}: Whether combining tasks destroys the result of previous tasks. For example, if Task A outputs a file and Task B moves it to a different location, Task B destroys the observable output of Task A, making it impossible to evaluate if Task A was successful after all tasks are completed. \\
\\
Output Format: Strictly JSON, e.g., \texttt{\{ "authenticity": 2, "rationality": 4, "evaluability": 5 \}} \\
\\
* Note: The GUI Agent has access to a network disk for file transfers. Cross-device transfers are implicitly included within the tasks. \\
\\
\textbf{Your Evaluation Target: }\{\{TASK CONTENT\}\} \\

\end{prompttable}

\subsection{Action Model}

After operating by the planner, tasks are decomposed into single-step GUI grounding actions executed on the screenshot of a single platform. To achieve the optimal grounding performance for all models, we keep all hyperparameters and prompts in the grounding step consistent with the models’ official grounding task settings. The prompt used is shown in Table~\ref{tab:grounding_prompt}.

\begin{prompttable}{tab:grounding_prompt}
{Prompt Template for Grounding Models}{Prompt Template for Grounding Models}
    \promptsection{Qwen3VL} \\
You are a helpful assistant. The user will give you an instruction, and you MUST left click on the corresponding UI element via tool call. If you are not sure about where to click, guess a most likely one.\\
\# Tools

You may call one or more functions to assist with the user query.

You are provided with function signatures within \textless tools\textgreater{} \textless /tools\textgreater{} XML tags: \\
\textless tools\textgreater{}\\
\{ "type": "function", "function": \{ "name": "computer\_use", "description": "Use a mouse to interact with a computer.\textbackslash n%
* The screen's resolution is 1000x1000.\textbackslash n%
* Make sure to click any buttons, links, icons, etc with the cursor tip in the center of the element.\textbackslash n%
* You can only use the left\_click action to interact with the computer.", "parameters": \{ "properties": \{ "action": \{ "description": "The action to perform. The available actions are:\textbackslash n%
* `left\_click`: Click the left mouse button with coordinate (x, y).", "enum": ["left\_click"], "type": "string" \}, "coordinate": \{ "description": "(x, y): The x (pixels from the left edge) and y (pixels from the top edge) coordinates to move the mouse to. Required only by `action=left\_click`.", "type": "array" \} \}, "required": ["action"], "type": "object" \} \} \} \\
\textless /tools\textgreater{}

For each function call, return a json object with function name and arguments within \textless tool\_call\textgreater{}\textless /tool\_call\textgreater{} XML tags:\\
\textless tool\_call\textgreater{}\\
\{ "name": \textless function-name\textgreater{}, "arguments": \textless args-json-object\textgreater{} \}\\
\textless /tool\_call\textgreater{}
\\
\promptsection{MAI-UI} \\
You are a GUI grounding agent. \\
\#\# Task\\
Given a screenshot and the user's grounding instruction. Your task is to accurately locate a UI element based on the user's instructions.\\
First, you should carefully examine the screenshot and analyze the user's instructions,  translate the user's instruction into a effective reasoning process, and then provide the final coordinate.\\
\#\# Output Format\\
Return a json object with a reasoning process in \textless grounding\_think\textgreater{}\textless /grounding\_think\textgreater{} tags, a [x,y] format coordinate within \textless answer\textgreater{} \textless /answer\textgreater{} XML tags:\\
\textless grounding\_think\textgreater{}...\textless /grounding\_think\textgreater{}\\
\textless answer\textgreater{}\\
\{"coordinate": [x,y]\}\\
\textless /answer\textgreater{}\\
\promptsection{HOLO2}\\
Localize an element on the GUI image according to the provided target and output a click position.\\
\ \ \ \ * You must output a valid JSON following the format: \{ClickCoordinates.model\_json\_schema()\}\\
Your target is:\\

\promptsection{UI-Venus}\\
Outline the position corresponding to the instruction: \{instruction\}. The output should be only [x1,y1,x2,y2].
\\
\promptsection{GUI-OWL for mobile}\\
You are a helpful assistant.\\

\# Tools\\

You may call one or more functions to assist with the user query.\\

You are provided with function signatures within \textless tools\textgreater{}\textless /tools\textgreater{} XML tags:\\
\textless tools\textgreater{}\\
\{\{"type": "function", "function": \{\{"name": "mobile\_use", "description": "Use a touchscreen to interact with a mobile device.\textbackslash n%
* This is an interface to a mobile device with touchscreen.\textbackslash n%
* The screen's resolution is \{width\}x\{height\}.\textbackslash n%
* Make sure to click any buttons, links, icons, etc with the cursor tip in the center of the element.\textbackslash n%
* Don't click boxes on their edges unless asked.", "parameters": \{\{"properties": \{\{"action": \{\{"description": "The action to perform. The available actions are:\textbackslash n%
* `click`: Click the point on the screen with coordinate (x, y).", "enum": ["click"], "type": "string"\}\}, "coordinate": \{\{"description": "(x, y): The x (pixels from the left edge) and y (pixels from the top edge) coordinates to click.", "type": "array"\}\}\}\}, "required": ["action", "coordinate"], "type": "object"\}\}\}\\
\textless /tools\textgreater{}\\

For each function call, return a json object with function name and arguments within \textless tool\_call\textgreater{}\textless /tool\_call\textgreater{} XML tags:\\
\textless tool\_call\textgreater{}\\
\{\{"name": \textless function-name\textgreater{}, "arguments": \textless args-json-object\textgreater{}\}\}\\
\textless /tool\_call\textgreater{}
\\
\promptsection{GUI-OWL for desktop}\\
You are a helpful assistant.\\

\# Tools\\

You may call one or more functions to assist with the user query.\\

You are provided with function signatures within \textless tools\textgreater{}\textless /tools\textgreater{} XML tags:\\
\textless tools\textgreater{}\\
\{\{"type": "function", "function": \{\{"name": "computer\_use", "description": "Use a mouse to interact with a computer.\textbackslash n%
* This is an interface to a desktop GUI.\textbackslash n%
* The screen's resolution is \{width\}x\{height\}.\textbackslash n%
* Make sure to click any buttons, links, icons, etc with the cursor tip in the center of the element.\textbackslash n%
* Don't click boxes on their edges unless asked.", "parameters": \{\{"properties": \{\{"action": \{\{"description": "The action to perform. The available actions are:\textbackslash n%
* `click`: Click the left mouse button at a specified (x, y) pixel coordinate on the screen.", "enum": ["click"], "type": "string"\}\}, "coordinate": \{\{"description": "(x, y): The x (pixels from the left edge) and y (pixels from the top edge) coordinates to click.", "type": "array"\}\}\}\}, "required": ["action", "coordinate"], "type": "object"\}\}\}\\
\textless /tools\textgreater{}\\

For each function call, return a json object with function name and arguments within \textless tool\_call\textgreater{}\textless /tool\_call\textgreater{} XML tags:\\
\textless tool\_call\textgreater{}\\
\{\{"name": \textless function-name\textgreater{}, "arguments": \textless args-json-object\textgreater{}\}\}\\
\textless /tool\_call\textgreater{}
\\
\promptsection{UI-Tars-1.5}\\
You are a GUI agent. You are given a task and your action history, with screenshots. You need to perform the next action to complete the task. \textbackslash n\textbackslash n%
\#\# Output Format\textbackslash n\textbackslash n%
Action: ...\textbackslash n\textbackslash n\textbackslash n%
\#\# Action Space\textbackslash n%
click(point='\textless point\textgreater{}x1 y1\textless /point\textgreater{}'')\textbackslash n\textbackslash n%
\#\# User Instruction \textbackslash n%
\ \{instruction\}
\end{prompttable}

\subsection{Environment Details}
\subsubsection{Infrastructure Details}
\label{Infrastructure Details}

All evaluations are conducted on two Linux machines, each equipped with 128GB of memory and properly configured with KVM to enable hardware acceleration. The evaluation is executed with a parallelism of 3. Although greater memory capacity enables more efficient parallel evaluation, it is not required for sequential execution. We provide configurable execution settings that allow users to adjust resource allocation according to their hardware configurations, including VM memory limits and parallelism. By setting the parallelism level to one, users can run the benchmark with a single virtual machine without requiring a large memory footprint.
The reproduction scripts, complete benchmark data, evaluation code, Docker orchestration files, model configurations, dashboard code, and all other resources required to reproduce the experiments will be publicly released. All models are deployed using vLLM, while Qwen3-VL-Plus is accessed via the Aliyun API as the planner model.

During evaluation, to reduce computational overhead, for devices that are irrelevant to a given task (i.e., the task does not require execution on these devices), we only capture the initial-state screenshots as visual inputs. When actions targeting such devices are issued, a ``device unavailable'' error is returned to the planner, preventing it from repeatedly attempting to operate on irrelevant devices. The network storage service web interface Alist is always enabled to ensure unrestricted access to relevant storage resources on task-related devices.

In terms of the execution mechanism of infrastructure layer, a pristine container is instantiated from a specified checkpoint for every task and discarded immediately after the task concludes. This ephemeral lifecycle ensures that \jarvisbench remains reproducible while providing a safe and isolated environment that effectively insulates the host machine.On the networking front, the layer leverages Docker's default mechanism where containers without explicit configurations automatically join the bridge network. This architecture seamlessly consolidates diverse heterogeneous containers (e.g., Android, Ubuntu, Windows) into a unified virtual subnet, facilitating inter-container communication.

\subsubsection{Environment Control Layer Details}
\label{Environment Control Layer Details}

In terms of perception, we define a Unified Observation Space incorporating both visual and structural information. This space consists of RGB screenshots, representing the current screen state, and the Accessibility Tree, describing the hierarchical relationships of UI widgets. The layer exposes standardized interfaces to acquire this data. Regarding implementation, for Ubuntu and Windows platforms, we adapted the Server-Controller architecture from OSWorld\citep{osworld}, streamlining the underlying codebase for improved efficiency. For the Android platform, we utilize ADB commands to control devices, abstracting these operations into unified interfaces.

We define a Unified Action Space compatible with both Desktop and Android platforms. Internally, this layer maps abstract actions to platform-specific execution primitives: employing PyAutoGUI for desktop control and ADB commands for Android devices. Externally, it exposes a standardized execute\_action interface to the upper layers. Our action space is shown in Table \ref{tab:action_space}. It encompasses not only basic interactive operations but also specific meta-controls, such as WAIT, FAIL, and DONE.
\input{tables/action_space}

\subsubsection{Evaluation Execution Layer Details}
\label{Dashboard Details}

\begin{figure}[!htbp]
    \centering
    \includegraphics[width=.85\linewidth]{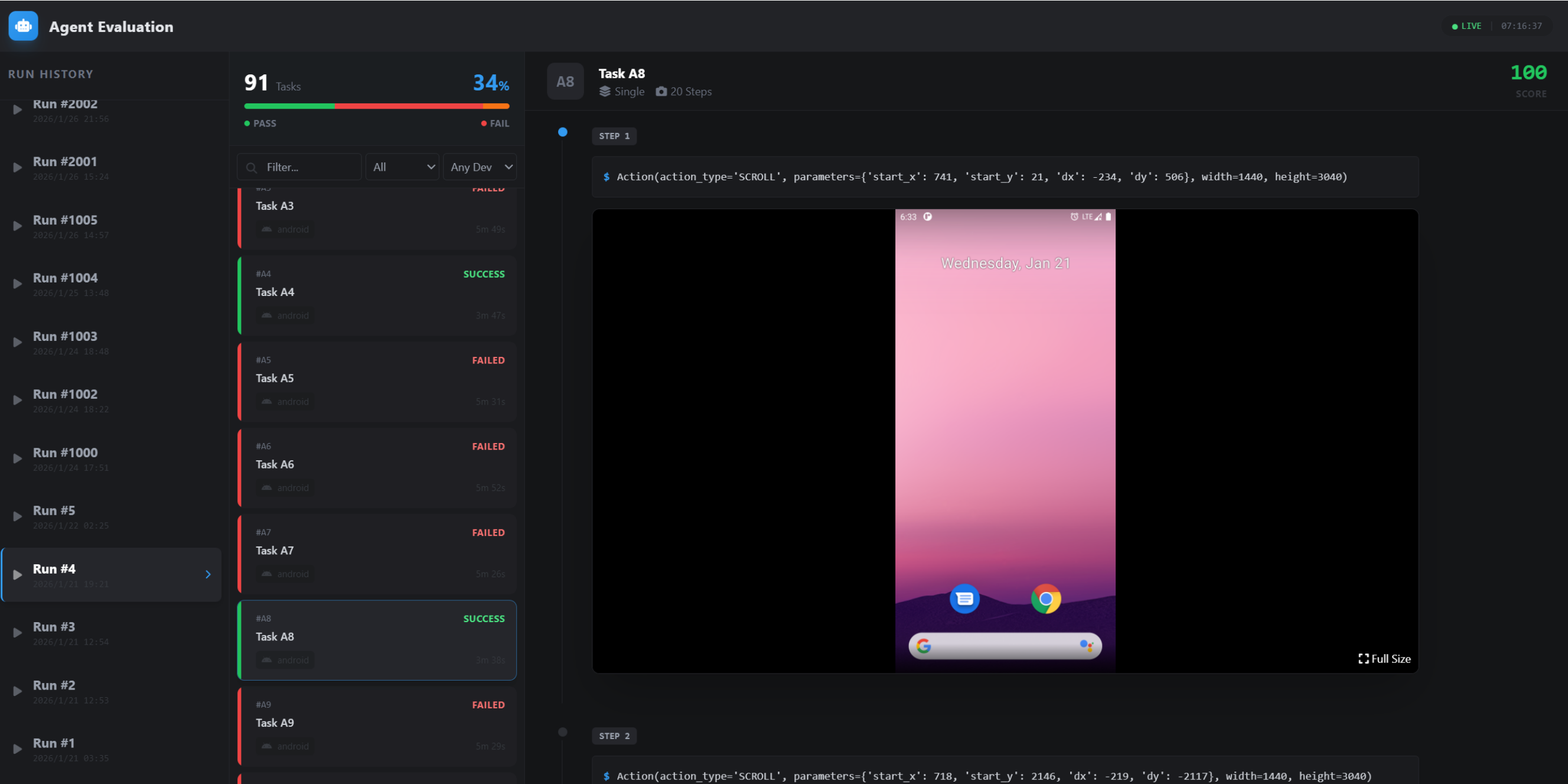}
    \caption{The Dashboard Interface}
    \label{fig:dashboard}
\end{figure}

To achieve high-throughput concurrent evaluation, we incorporate a Thread Pool Scheduling mechanism and an Asynchronous Environment Pre-warming strategy. Thread Pool Scheduling maximizes the utilization of the host's multi-core computing resources for efficient parallel execution. Meanwhile, Asynchronous Environment Pre-warming allows for the proactive initialization of container resources for subsequent tasks in the background, concurrent with the execution of ongoing tasks. This strategy effectively eliminates the idle latency associated with container cold starts (which typically accounts for 30\%-50\% of the total evaluation cycle).

To facilitate the tracking, analysis, and visualization of experimental results, we developed a web-based Agent Evaluation Dashboard, as shown in Figure \ref{fig:dashboard}. This tool provides an intuitive interactive interface that enables researchers and developers to monitor evaluation runs in real time and analyze the results.

To initiate the backend server, execute the command \texttt{python start\_dashboard.py}. 
The server listens on port 8000 by default, which can be customized using the \texttt{-{}-port} argument. 
Once the server is running, the dashboard is accessible via a web browser at \texttt{http://localhost:8000/viewer.html}.

\subsubsection{Configuration File Details}
\label{Environment Configuration Details}
\begin{figure*}[!ht]     \centering     \includegraphics[width=.85\linewidth]{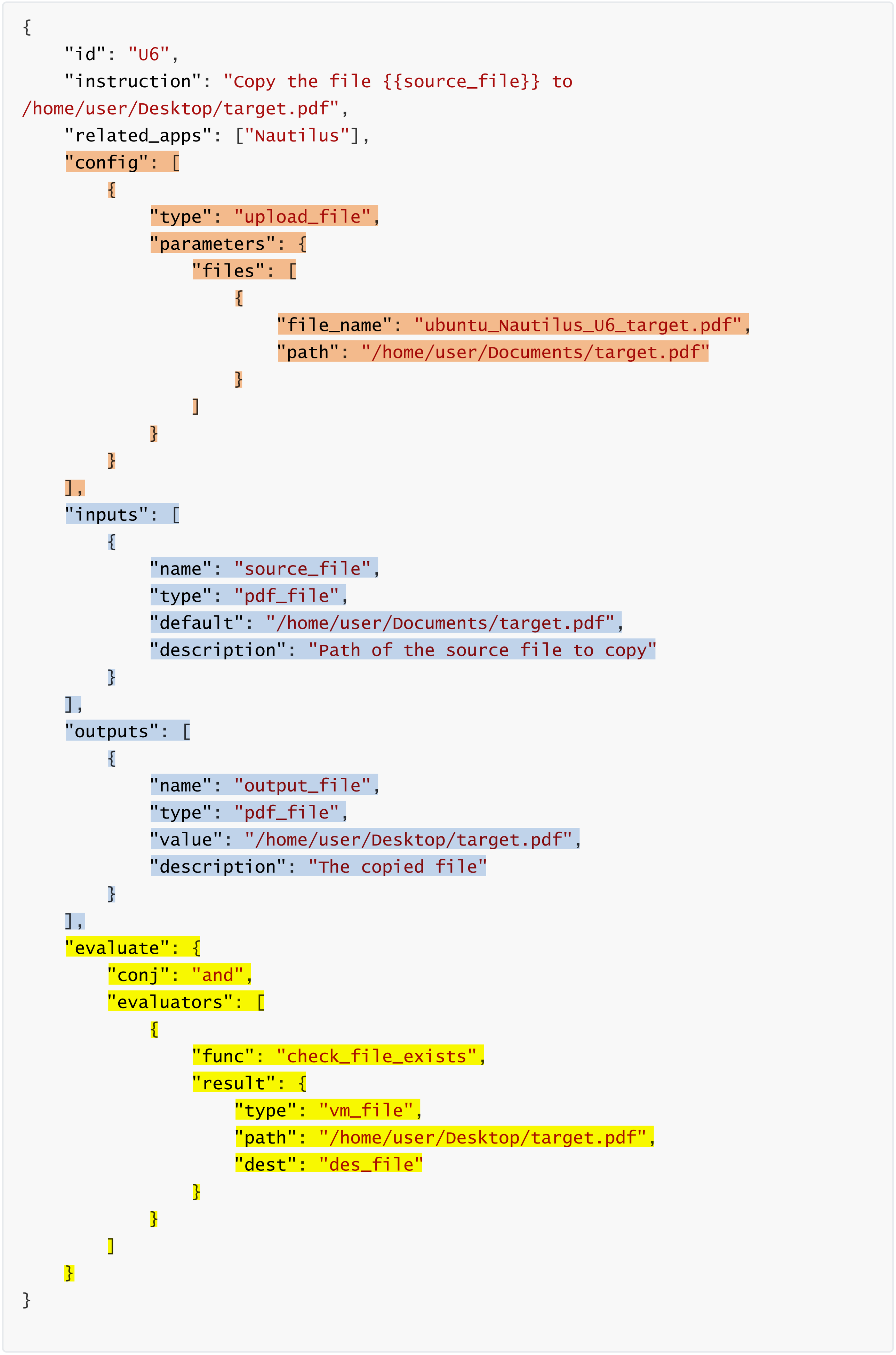}     \caption{The Single Task JSON. A subtask configuration generally comprises three primary segments: 
the \texttt{config} field (\textbf{highlighted in orange}), designed to establish execution \textbf{prerequisites}; 
the input and output fields (\textbf{highlighted in blue}), utilized for \textbf{dynamic data composition}; 
and the evaluation field (\textbf{highlighted in yellow}), which defines the specific \textbf{evaluation rules}.}     \label{fig:configuration} \end{figure*}

Our dataset is encapsulated within a single JSON file. This file contains a \texttt{tasks} field, which comprises a list of dictionaries, where each item represents an individual task configuration, as shown in Figure \ref{fig:configuration}.

Each single task configuration entry primarily consists of the following fields: \texttt{id}, \texttt{devices}, \texttt{instruction}, and \texttt{formatted\_subtasks}. 
The \texttt{formatted\_subtasks} field contains a list of configurations for all associated subtasks, where each item includes a specific \texttt{config} field (highlighted with orange in Figure \ref{fig:configuration}) and an \texttt{evaluate} field (highlighted with yellow in Figure \ref{fig:configuration}). Certain tasks incorporate input and output fields to facilitate dynamic data composition (highlighted with blue in Figure \ref{fig:configuration}). Specifically, input fields serve as template variables, which can be referenced within the \texttt{instruction} and \texttt{evaluate} fields using the \texttt{{{}}} placeholder syntax.

\subsection{Typed Automatic Task Composition Details}

\label{alg:task_stitching}

\begin{algorithm}[!ht]
\caption{Typed Automatic Task Composition}
\begin{algorithmic}
\STATE {\bfseries Input:} task set $\mathcal{T}$, type system $(\mathbb{T}, \preceq)$, maximum number of tasks $K$
\STATE {\bfseries Output:} composed task graph $\mathcal{G} = (\mathcal{V}, \mathcal{E})$

\STATE Sample an initial task instance $v_0$ from $\mathcal{T}$
\STATE Initialize $\mathcal{V} \leftarrow \{v_0\}$, $\mathcal{E} \leftarrow \emptyset$

\WHILE{$|\mathcal{V}| < K$}
    \STATE Identify unfilled input slots $\mathcal{I}_{open}$ from tasks in $\mathcal{V}$
    \STATE Identify available output slots $\mathcal{O}_{avail}$ from tasks in $\mathcal{V}$

    \STATE Sample a slot $s$ from $\mathcal{I}_{open} \cup \mathcal{O}_{avail}$

    \STATE Find a task instance $v \in \mathcal{T}$ with a compatible slot $s'$ such that either
    \STATE \hspace{1em} (i) $s' \in \mathcal{O}_v$, $s \in \mathcal{I}$, and $\tau_{s'} \preceq \tau_{s}$, or
    \STATE \hspace{1em} (ii) $s \in \mathcal{O}$, $s' \in \mathcal{I}_v$, and $\tau_{s} \preceq \tau_{s'}$

    \IF{no compatible task instance exists}
        \STATE \textbf{break}
    \ENDIF

    \STATE Add $v$ to $\mathcal{V}$
    \STATE Add a directed edge connecting $s$ and $s'$ to $\mathcal{E}$

    \IF{$P_v \neq P_{u}$ for the connected tasks}
        \STATE Insert a platform transfer task to ensure cross-platform execution
    \ENDIF
\ENDWHILE

\STATE Return $\mathcal{G} = (\mathcal{V}, \mathcal{E})$
\end{algorithmic}
\end{algorithm}
In this appendix, we provide additional details on the task composition procedure used in our experiments. The overall construction process is formalized in Algorithm~\ref{alg:task_stitching}, which describes a typed automatic task composition method for building executable task graphs from a heterogeneous task library. 

The algorithm takes as input a task set $\mathcal{T}$, a partially ordered type system $(\mathbb{T}, \preceq)$, and a maximum graph size $K$. Starting from a randomly sampled initial task, the procedure incrementally grows a directed task graph by iteratively connecting compatible input and output slots. Slot compatibility is determined by the type preorder $\preceq$, ensuring that data flows only from more specific types to more general ones, or vice versa, depending on slot direction.

At each iteration, the algorithm samples either an unfilled input slot or an available output slot from the current graph, and searches for a task instance in $\mathcal{T}$ that provides a compatible counterpart. If no such task exists, graph construction terminates early. This stochastic expansion strategy allows the method to generate diverse task graphs while maintaining type safety.

In addition, Algorithm~\ref{alg:task_stitching} explicitly handles cross-platform execution constraints. When two connected tasks are associated with different execution platforms, an auxiliary platform transfer task is inserted to preserve executability. This design enables seamless composition across heterogeneous environments without violating platform assumptions.

Overall, the proposed algorithm provides a flexible and principled mechanism for automatically synthesizing typed task graphs, which serves as the backbone for the experimental evaluations presented in the main paper.

\subsection{Task detailed statistics}
\label{sec:task_detailed_statistics}

\begin{figure}[!ht]
    \centering
    \includegraphics[height=0.5\linewidth]{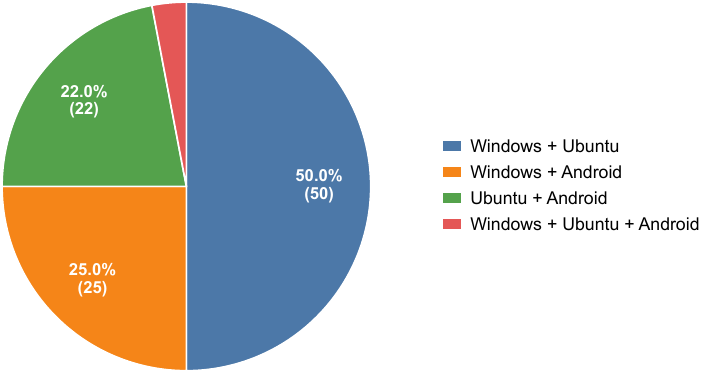}
    \caption{Task Distribution over device combinations.}
    \label{fig:device_combinations}
\end{figure}

\begin{figure}[!ht]
    \centering
    \includegraphics[height=0.5\linewidth]{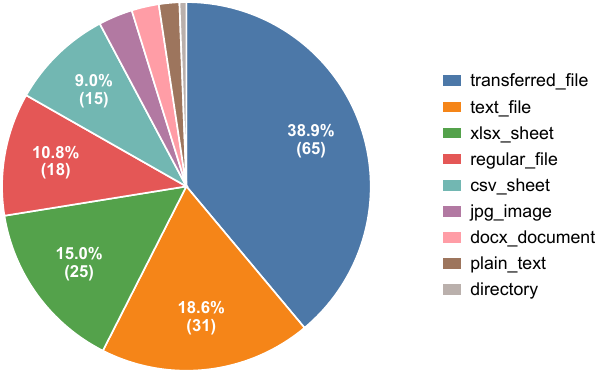}
    \caption{Task Distribution over slot type.}
    \label{fig:slot_type}
\end{figure}

\jarvisbench covers a diverse set of device combinations and multiple source object types, as illustrated in Figure~\ref{fig:device_combinations} and Figure~\ref{fig:slot_type}, respectively. 

Notably, the \texttt{transferred\_file} slot type represents files that need to be transferred across platforms in auxiliary tasks, allowing arbitrary file formats. Since \jarvisbench contains a large number of cross-platform tasks, this slot type accounts for a relatively large proportion, which is expected.

\subsection{Experiment Details}
\label{sec:experiment_details}

To further investigate whether the observed bottleneck is specific to our choice of planner (Qwen3-VL-Plus), we additionally tested Llama 4 Maverick and Kimi K2.6 under the same experimental setting. We selected these models because they represent recent publicly available multimodal large language models with strong vision-language reasoning and agentic capabilities. In particular, Llama 4 Maverick and Kimi K2.6 demonstrate competitive performance on widely used multimodal and agent-oriented benchmarks, including MMMU-style visual reasoning and OSWorld-style computer interaction evaluations. Therefore, they provide diverse planner backbones for examining whether the observed cross-device bottleneck generalizes beyond Qwen3-VL-Plus.

Under our current experimental protocol, Llama 4 Maverick showed limited instruction-following reliability: it frequently failed to produce outputs in the required action format and, consequently, achieved near-zero evaluation scores across all tasks due to its inability to consistently generate executable actions. 

Kimi K2.6 achieved moderately higher performance than Qwen3-VL-Plus on both atomic and compositional tasks. Nevertheless, its performance on cross-device tasks remained limited. As shown in Table \ref{tab:additional_planners}, the Multi-Overall score only increased from 8.0 to 11.3. These results show that replacing the planner may improve the performance of the complete system, but does not eliminate the substantial degradation on cross-device compositional tasks. This provides further evidence that the performance bottleneck is broadly present across the models rather than being specific to Qwen3-VL-Plus.

\input{tables/additional_planner}

To provide a more rigorous statistical assessment of the model performances, we report the raw counts of successful tasks alongside their percentages. Furthermore, we calculated the 95\% bootstrap confidence intervals over the tasks. These detailed statistics are presented in Table~\ref{tab:confidence_intervals}.

\input{tables/confidence_interval}

\section{Empirical Evidence on the Prevalence of Cross-Device Workflows}
\label{sec:empirical-cross-device}

To assess the prevalence of cross-device workflows in real-world settings and to further support the practical value of cross-device GUI research, we summarize empirical evidence from prior (M)LLM research and HCI studies. From the perspective of (M)LLM applications, the deployment scope has steadily expanded: from early dialogue systems \cite{liu2020towards}, to terminal environments \cite{cheng2026terminalworldscalingterminalagentenvironments,cheng2026mem2evolve}, to today's GUI agents and embodied intelligence \cite{lan2026peap}, and most recently to chemistry and other fundamental sciences \cite{gao2025chemical}. This trajectory indicates that as (M)LLM capabilities continue to improve, users are delegating an ever-growing range of tasks to them. Cross-device workflows, which HCI studies have shown to be a common pattern of everyday work, are therefore a natural next target: there is both the potential and the demand for (M)LLMs to take them over.

Although large-scale logs of cross-device behavior are difficult to obtain due to privacy constraints, the HCI evidence on this point is consistent: cross-device workflows are common and practically relevant. First, multi-device ownership is widespread: Brudy et al.~\cite{brudy2018overview} report that 67.5\% of participants owned two or more devices, with an average of 1.9 devices per participant (SD = 0.7). Second, users explicitly value continuity across devices. Raptis et al.~\cite{raptis2016continuity} conducted a large-scale online study analyzing 1,603 valid user reviews of Apple's continuity features and found strong demand for cross-device capabilities; for example, 23.3\% of reviews emphasized breaking device barriers, while 14.3\% asked for broader support across more applications. The same study further shows that users often switch devices because a task cannot be completed on the current device. Third, cross-device use is routine in everyday work and leisure. Majrashi et al.~\cite{majrashi2021crossdevice} show that people regularly combine phones, computers, tablets, and other devices in daily activities, while Jokela et al.~\cite{jokela2015diary} report that 37\% of recorded multi-device cases involved sequential use of multiple devices within a single task. Taken together, these findings suggest that cross-device workflows are not rare or artificial scenarios, but common practices in everyday device use, thereby providing additional empirical support for the real-world relevance of cross-device GUI research.

\section{Future Extension}
\label{sec:appendix_future_work}

The current release of \jarvisbench establishes a robust foundational scale, providing strong statistical power for our reported results. Unlike prior work such as CRAB, which contains 18 manually designed cross-device tasks, our benchmark comprises 150 composite workflows containing 442 platform-specific subtasks and 65 file-transfer auxiliary tasks (averaging approximately three subtasks per workflow). These span dozens of applications across multiple domains, creating a rich and complex state-transition space. Moving forward, we view this scale as a strong starting point. By leveraging our structured input--output modeling of atomic tasks, future extensions will automatically recompose and sample an even broader range of cross-device workflows to ensure continuous scaling of task diversity.

As we scale, future extensions will also aim to broaden the physical and systemic scenario coverage, which is currently constrained by necessary experimental trade-offs. To ensure reproducibility and manage evaluation costs, our current device combinations focus on Android, Windows, and Ubuntu (excluding macOS and iOS due to restrictive virtualization licenses). We also cap trajectory lengths at 50 steps to avoid the sharply increased inference costs of ultra-long loop-based tasks, and utilize a private AList-based network drive for file transfers rather than simulating physical USBs or unstable Bluetooth protocol stacks. Future iterations of \jarvisbench aim to bridge these gaps by exploring advanced simulation environments for closed ecosystems, optimized inference frameworks to support ultra-long horizon planning, and virtualized hardware protocols to capture a more complete spectrum of real-world interactions.

Finally, expanding the diversity of target user scenarios remains a crucial priority. While the current iteration primarily focuses on generalized mainstream workflows, GUI agents hold immense potential to assist elderly and disabled users who may face physical or cognitive barriers in multi-platform environments. Developing accessibility-oriented cross-device tasks—such as automated text-to-speech synchronization across devices, simplified remote health monitoring setups, or cross-platform accessibility setting configurations—is an important direction for future extension. We plan to collaborate with HCI researchers and target user groups to systematically incorporate these socially impactful workflows into future versions of the benchmark.

\section{Failure Modes and Case Studies}

\subsection{Failure Modes}
\label{sec:failure_mode}

To better understand why agents failed to complete compositional tasks, we conducted a further analysis. Our failure analysis focuses on observable action trajectories rather than the underlying causes because the same error may have several possible explanations. For example, an agent may move to the target device before the file upload is complete. This may happen because it forgot the target path, clicked the wrong GUI element, incorrectly interpreted the interface feedback and assumed that the upload had finished or other factors. Since these different causes can lead to the same observed behavior, it is difficult to determine a unique cause from the incorrect actions alone. Such attribution may therefore be subjective and unreliable. 

\begin{figure}[!h]
    \centering
    \includegraphics[height=.5\linewidth]{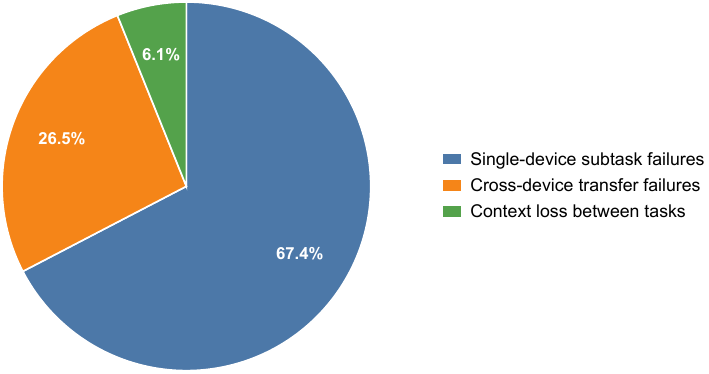}
    \caption{Failure mode of HOLO2 during compositional tasks.}
    \label{fig:failure_mode}
\end{figure}

Our preliminary analysis of HOLO2 trajectories on cross-device tasks reveals three major failure modes, as shown in Fig.~\ref{fig:failure_mode}: (a) failures stemming from underlying single-device subtasks; (b) failures in cross-device transfer mechanisms; and (c) failures caused by loss of context between tasks. Additionally, we manually examined representative failure cases from the other evaluated models and found that they exhibit similar failure patterns.

\subsection{Case Study}
\label{appendix_case_study}
Current GUI agents struggle with the majority of cross-device tasks, while achieving a non-trivial success rate on single-device atomic tasks, particularly on Ubuntu. We first report common failure cases observed in single-device task execution, and then analyze errors arising in multi-device settings.
\subsubsection{Atomic Tasks}

\begin{figure*}[!ht]
    \centering
    \includegraphics[width=.85\linewidth]{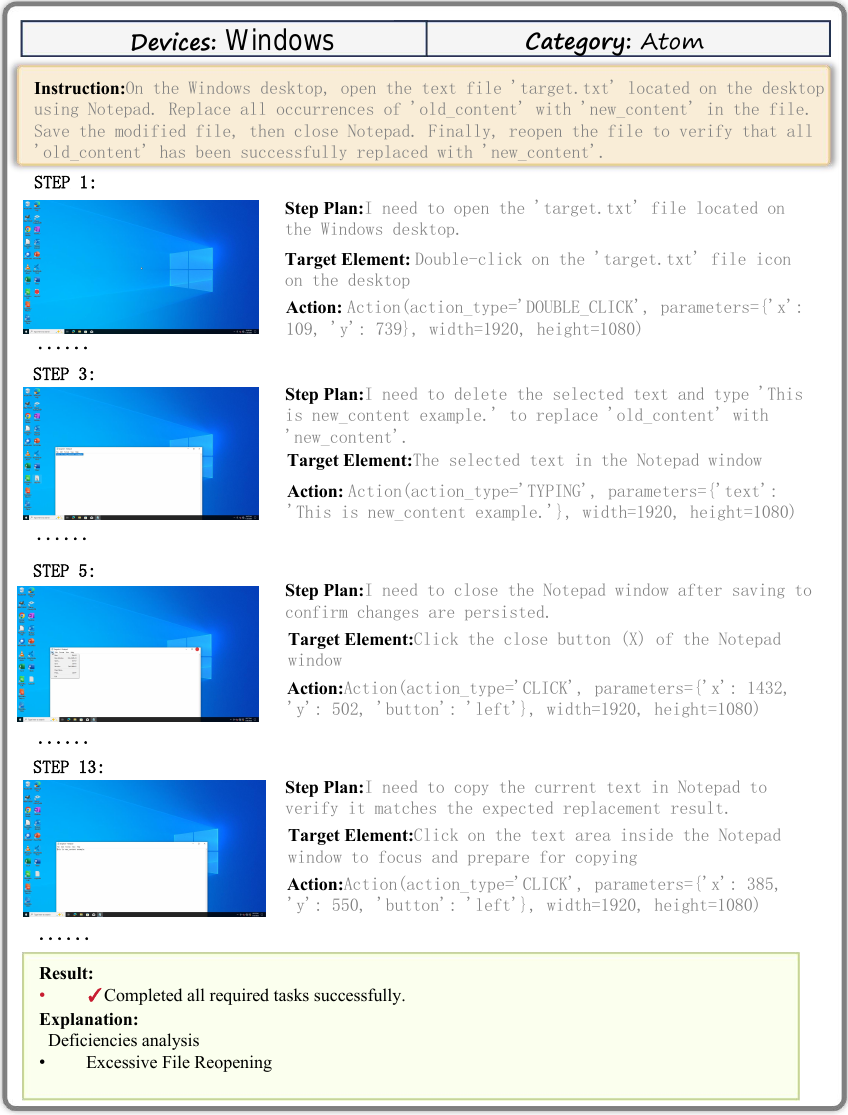}
    \caption{Successful Case of an Atomic Task}
    \label{fig:atom_success}
\end{figure*}

\paragraph{Successful Case}
Figure~\ref{fig:atom_success} illustrates a successful case of an atomic task. In this example, the agent successfully opens Notepad on Windows, correctly recognizes the text contained in the \texttt{.txt} file, and accurately replaces the target text with the specified content according to the given instruction.

\begin{figure*}[!ht]
    \centering
    \includegraphics[width=.85\linewidth]{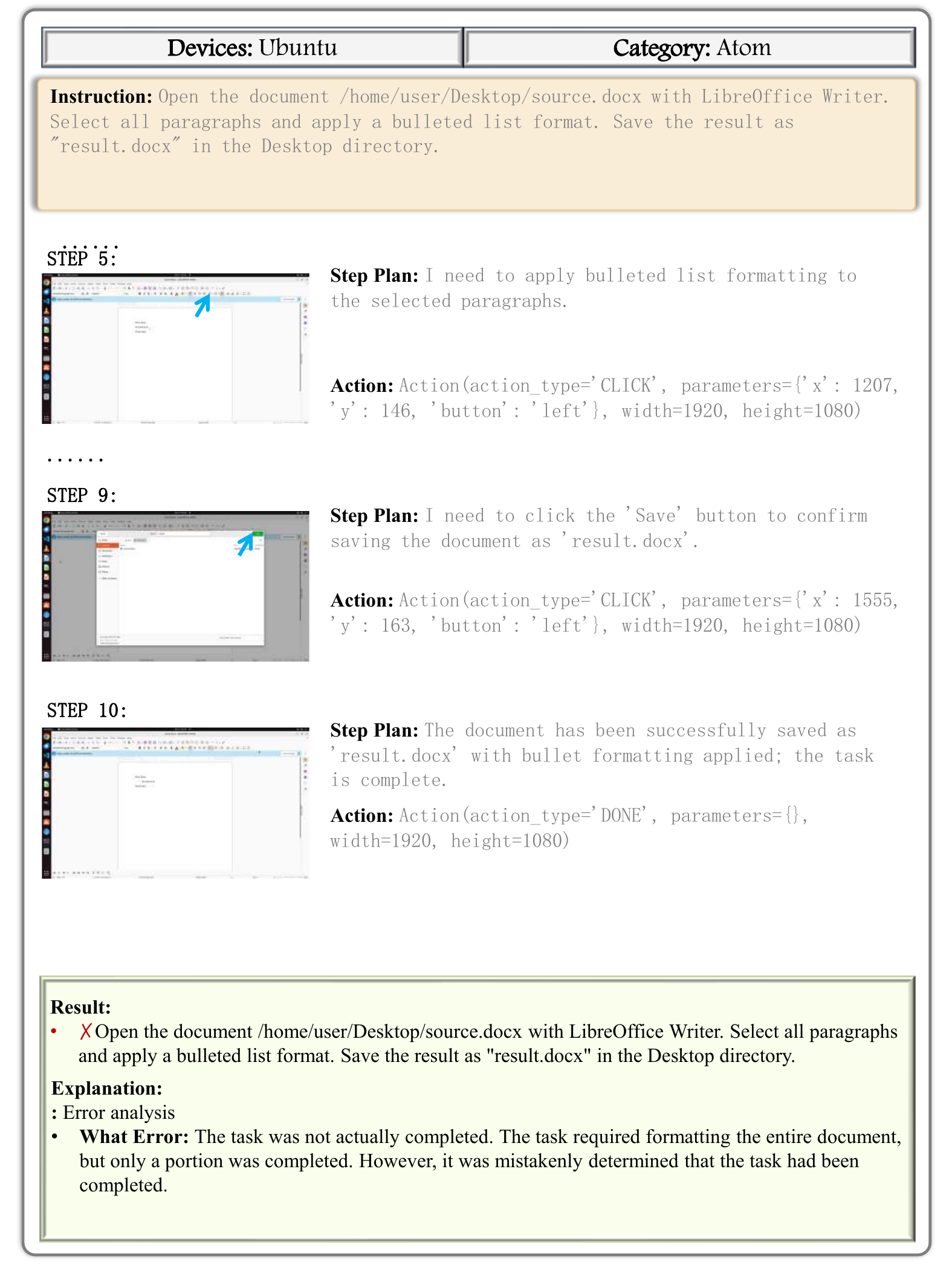}
    \caption{Failure Case of an Atomic Task Due to an Incorrect Planner Completion Judgment}
    \label{fig:atom_planner_error1}
\end{figure*}

\paragraph{Failure Case}
Figure~\ref{fig:atom_planner_error1} illustrates a representative failure case. In this example, the agent successfully opens LibreOffice Writer, selects a portion of the text, and correctly applies a bulleted list format to the selected lines. However, the task requires applying the formatting to all lines, which is not fully accomplished. The planner nevertheless incorrectly judges the task as completed.

\begin{figure*}[!ht]
    \centering
    \includegraphics[width=.85\linewidth]{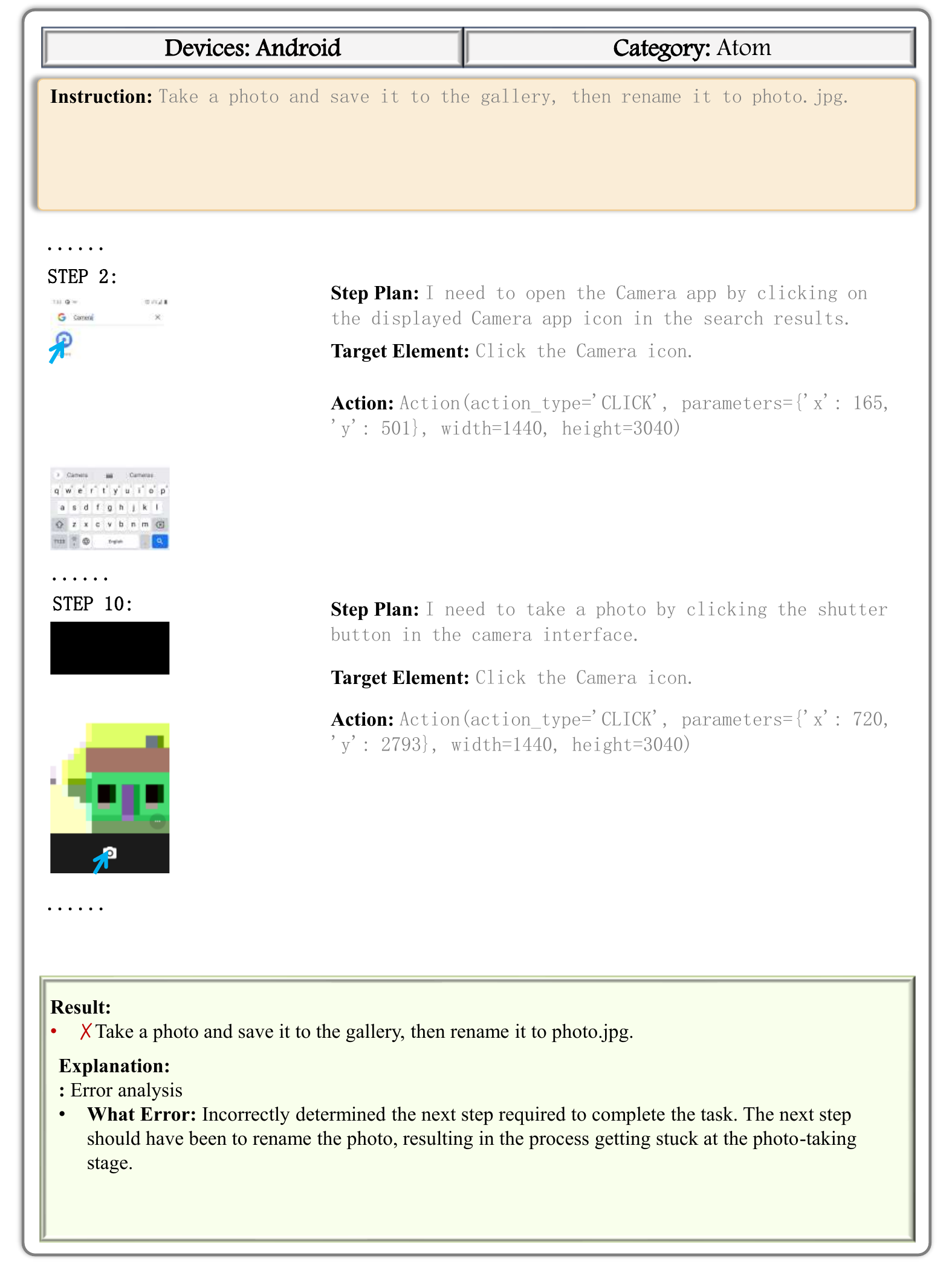}
    \caption{Failure Case of an Atomic Task Due to the Planner’s Failure to Identify the Next Action}
    \label{fig:atom_planner_error2}
\end{figure*}

Figure~\ref{fig:atom_planner_error2} illustrates another type of failure. In this case, the agent successfully opens the camera application and captures a photo. However, the task further requires renaming the captured image to a specified filename, a step that is not completed because the planner is unable to determine the appropriate next action.

\begin{figure*}[!ht]
    \centering
    \includegraphics[width=.85\linewidth]{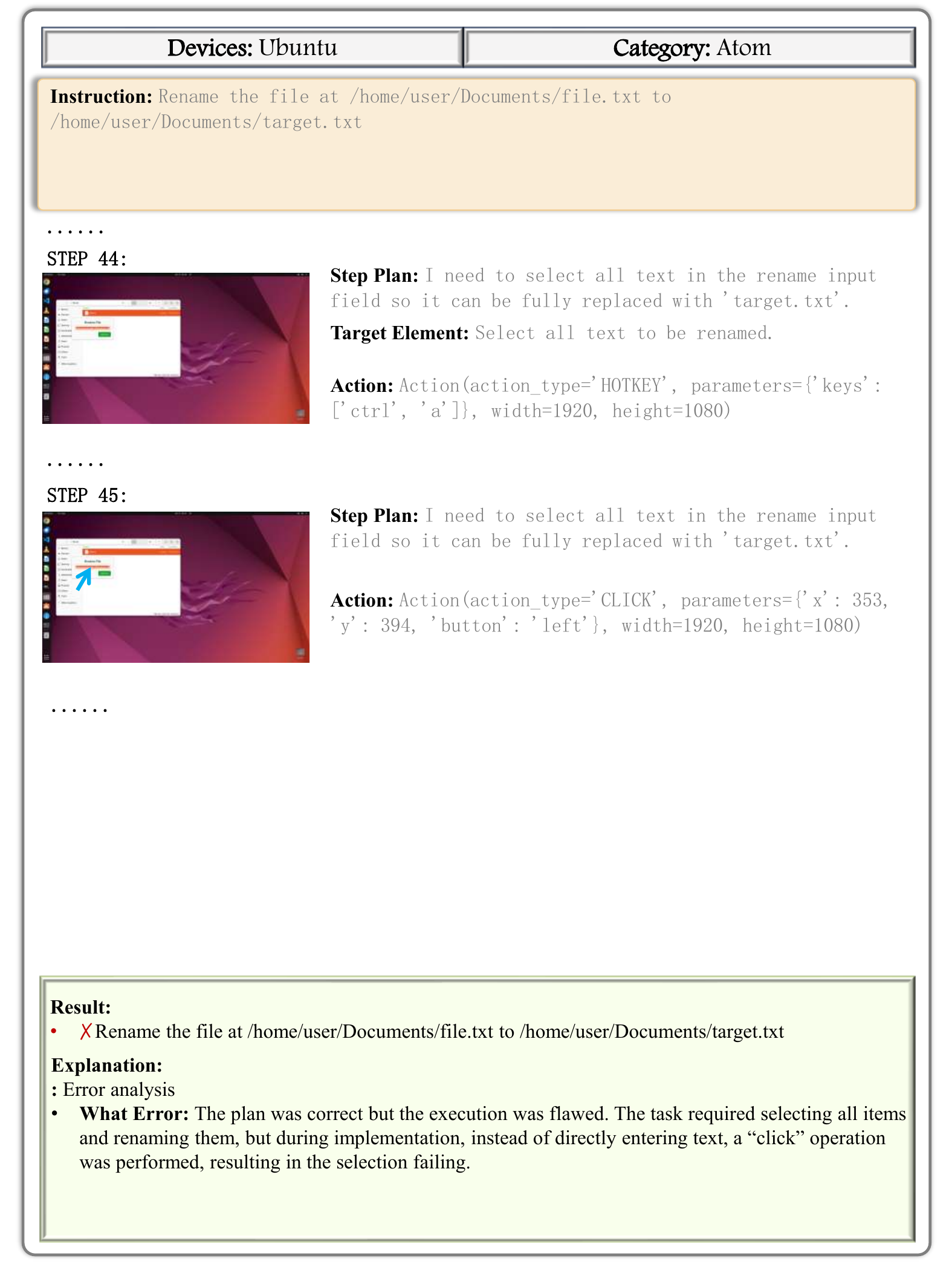}
    \caption{Failure Case of an Atomic Task Due to the Planner’s Failure to Predict the Atomic Action}
    \label{fig:atom_planner_error3}
\end{figure*}

Figure~\ref{fig:atom_planner_error3} illustrates a failure case in which the planner fails to predict the required atomic action, preventing the agent from correctly completing the file renaming operation.

\subsubsection{Compositional Tasks}

\begin{figure*}[!ht]
    \centering
    \includegraphics[width=.85\linewidth]{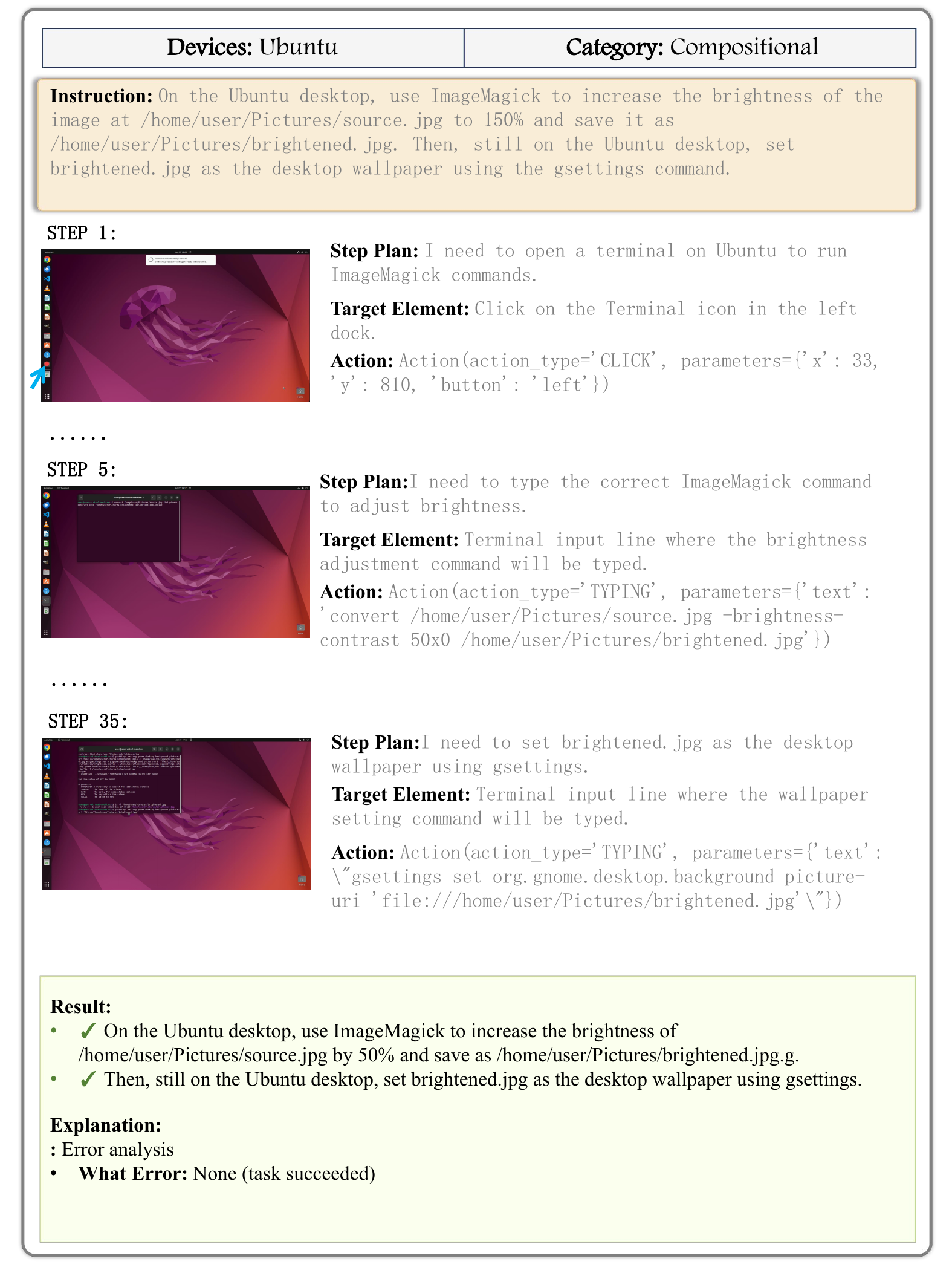}
    \caption{A Successful Example of a Compositional Task with Dependencies on a Single Device}
    \label{fig:comp_successful_1}
\end{figure*}

\begin{figure*}[!ht]
    \centering
    \includegraphics[width=.85\linewidth]{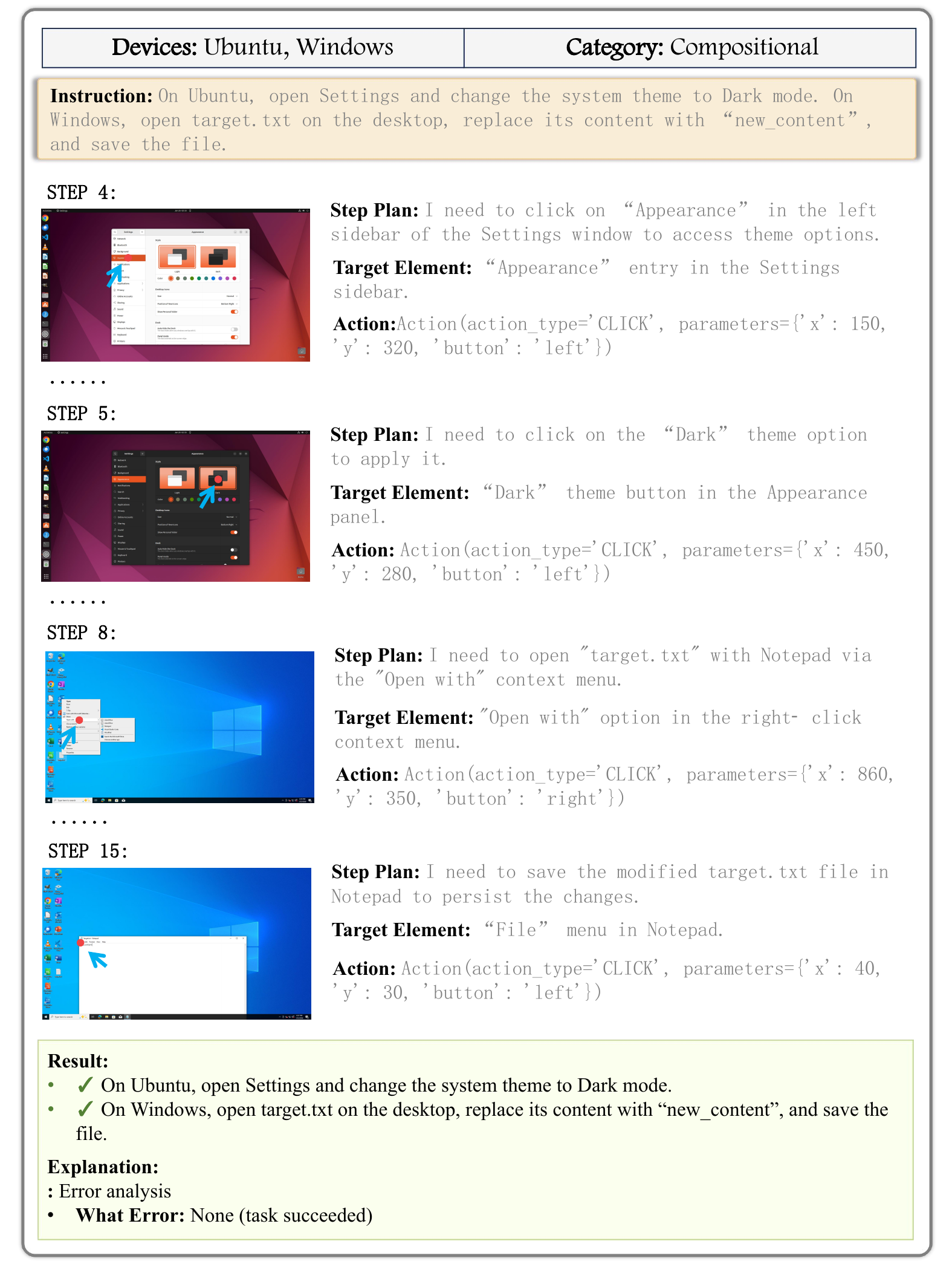}
    \caption{A Successful Example of a Compositional Task without Dependencies on Multiple Devices}
    \label{fig:comp_successful_2}
\end{figure*}

\paragraph{Successful Case}
Two illustrative examples demonstrate the system’s capability to execute compositional tasks across different device contexts. The first example (Figure~\ref{fig:comp_successful_1}) showcases a dependency-free workflow executed entirely on an Ubuntu desktop, where the system first increases the brightness of a source image using ImageMagick and subsequently applies the processed image as the desktop wallpaper via gsettings. The second example (Figure~\ref{fig:comp_successful_2}) highlights cross-device task execution: the system switches the Ubuntu environment to Dark mode while concurrently modifying the contents of target.txt on a Windows machine. Together, these examples illustrate reliable task coordination in both single-device and multi-device settings.

\begin{figure*}[!ht]
    \centering
    \includegraphics[width=.85\linewidth]{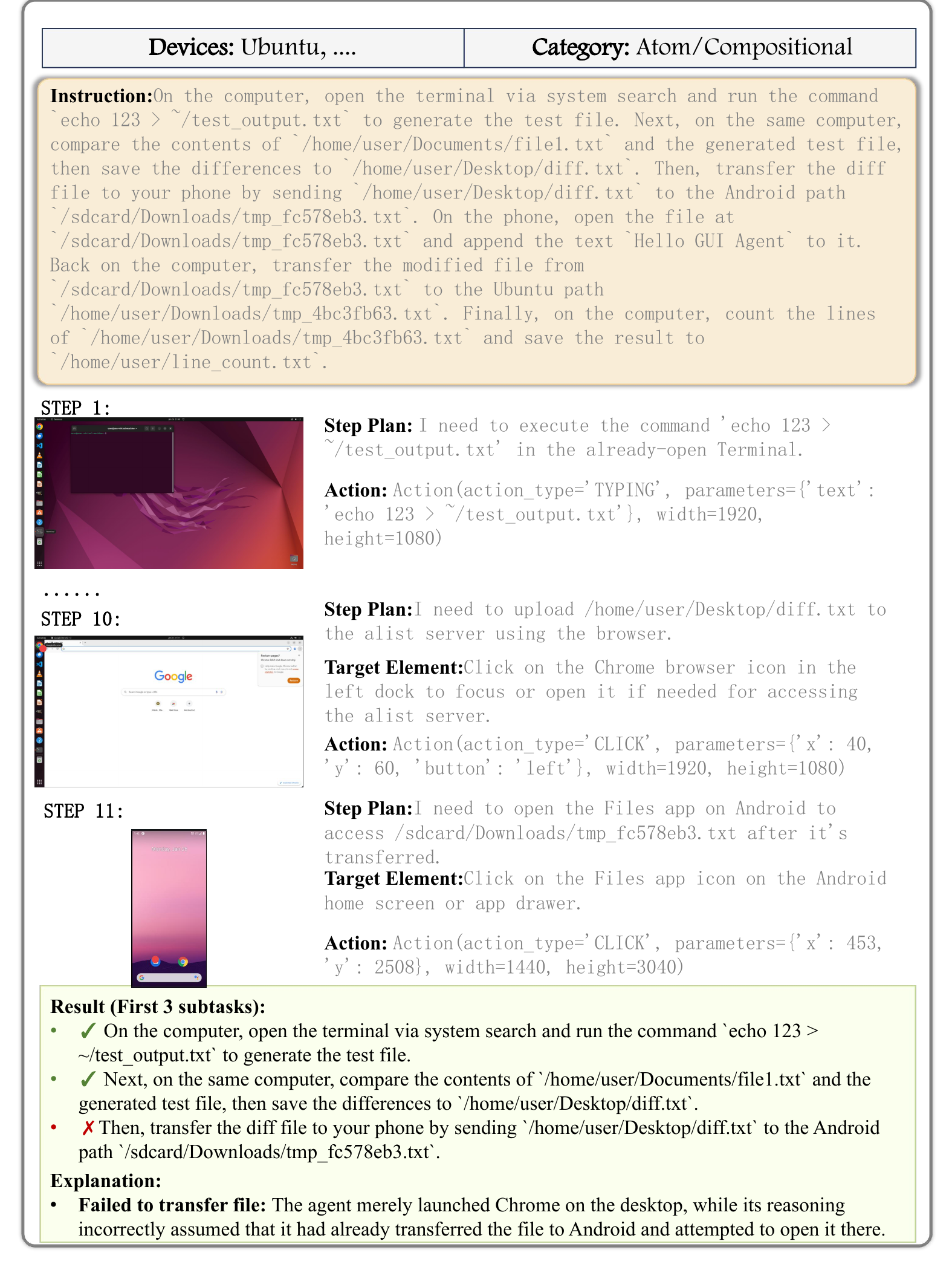}
    \caption{Failure Case of an Compositional Task Due to Failed to Transfer File.}
    \label{fig:comp_fail_1}
\end{figure*}

\paragraph{Failure Case}
In the case shown in Figure~\ref{fig:comp_fail_1}, the agent only opened Chrome and did nothing further, while its reasoning indicated that it had already transferred the file to Android and attempted to open it there.

%% file: tables/action_space.tex
\begin{table}[htbp]

    \centering
    \small
    \begin{tabular}{ll}
    \toprule
    \textbf{Action} & \textbf{Description} \\
    \midrule
    \texttt{MOVE\_TO}$(x, y)$ & Move cursor to coordinate. \\
    \texttt{CLICK}$(\dots)$ & Click current/target pos with options. \\
    \texttt{MOUSE\_DOWN/UP}$(btn)$ & Press/Release mouse button. \\
    \texttt{RIGHT\_CLICK}$(x, y)$ & Right-click at position. \\
    \texttt{DOUBLE\_CLICK}$(x, y)$ & Double left-click at position. \\
    \texttt{DRAG\_TO}$(\dots)$ & Drag from start to target. \\
    \texttt{SCROLL}$(dx, dy)$ & Scroll mouse wheel. \\
    \texttt{SWIPE}$(\dots)$ & Touch swipe by offset (Mobile). \\
    \texttt{LONG\_PRESS}$(x, y, t)$ & Long press for $t$ ms. \\
    \texttt{TYPING}$(text)$ & Input text string. \\
    \texttt{KEY\_DOWN/UP}$(key)$ & Press/Release specific key. \\
    \texttt{HOTKEY}$(keys)$ & Press key combination. \\
    \texttt{WAIT}$(sec)$ & Pause execution. \\
    \texttt{FAIL/DONE}() & End task (fail/success). \\
    \texttt{PYAUTOGUI}$(script)$ & Exec raw Python script. \\
    \bottomrule
    \end{tabular}

        \caption{Action Space of \jarvisbench. CLICK has parameters x, y, button, nClicks; DRAG\_TO has parameters x, y, startX, startY; SWIPE has parameters dx, dy, x, y, duration.}
    \label{tab:action_space}
\end{table}

%% file: tables/additional_planner.tex
\begin{table*}[!ht]
    \centering
    \small

    \renewcommand\arraystretch{1.2} 
    
    \resizebox{1\linewidth}{!}{
        \begin{tabular}{l ccc c cccc}
            \toprule
            
            \multirow{2}{*}{\textbf{Model}} 
            & \multicolumn{4}{c}{\textbf{Atomic Tasks}} & \multicolumn{4}{c}{\textbf{Multi-Tasks}} \\
            
            \cmidrule(lr){2-5} \cmidrule(lr){6-9}
            
             & \multicolumn{1}{c}{Android} & \multicolumn{1}{c}{Windows} & \multicolumn{1}{c}{Ubuntu} & \multicolumn{1}{c}{Overall}
             & \multicolumn{1}{c}{SW} & \multicolumn{1}{c}{MI} & \multicolumn{1}{c}{MD} & \multicolumn{1}{c}{Overall} \\
            \midrule
            
            HOLO2 (Qwen3-VL-Plus) 
            & 41.7 & 15.8 & 60.7 & 42.4 
            & 16.0 & 6.0 & \best{2.0} & 8.0 \\

            HOLO2 (Kimi K2.6) 
            & \best{45.8} & \best{23.7} & \best{66.1} & \best{48.3} 
            & \best{22.0} & \best{10.0} & \best{2.0} & \best{11.3} \\

            \bottomrule
        \end{tabular}
    }

    \caption{
    \textbf{Performance comparison using different planner backbones.} 
    Experimental results testing alternative LLMs within the HOLO2 architecture to verify cross-device bottleneck generalizability.
    The best results are highlighted in \textbf{bold}.
    }
    \label{tab:additional_planners}
\end{table*}

%% file: tables/confidence_interval.tex
\begin{table*}[!ht]
    \centering
    \small

    \renewcommand\arraystretch{1.2} 
    
    \resizebox{1\linewidth}{!}{
        \begin{tabular}{l lll}
            \toprule
            \textbf{Model} & \textbf{Atomic task} & \textbf{Composite task} & \textbf{Composite subtask} \\
            \midrule
            
            UI-Venus 
            & 37.3\% [28.8\%, 45.8\%] (44/118) 
            & 1.3\% [0.0\%, 3.3\%] (2/150) 
            & 16.7\% [13.0\%, 20.8\%] (74/442) \\
            
            UI-TARS-1.5 
            & 37.3\% [28.8\%, 45.8\%] (44/118) 
            & 6.0\% [2.7\%, 10.0\%] (9/150) 
            & 18.8\% [14.5\%, 23.4\%] (83/442) \\
            
            MAI-UI 
            & 28.8\% [21.2\%, 37.3\%] (34/118) 
            & 4.0\% [1.3\%, 7.3\%] (6/150) 
            & 16.1\% [12.2\%, 20.2\%] (71/442) \\
            
            GUI-Owl 
            & 39.0\% [30.5\%, 48.3\%] (46/118) 
            & 7.3\% [3.3\%, 12.0\%] (11/150) 
            & 18.1\% [14.1\%, 22.6\%] (80/442) \\
            
            Qwen3-VL 
            & 35.6\% [27.1\%, 44.1\%] (42/118) 
            & 4.0\% [1.3\%, 7.3\%] (6/150) 
            & 18.1\% [14.3\%, 22.3\%] (80/442) \\
            
            HOLO2 
            & \best{42.4\%} [33.9\%, 51.7\%] (50/118) 
            & \best{8.0\%} [4.0\%, 12.7\%] (12/150) 
            & \best{19.0\%} [14.8\%, 23.6\%] (84/442) \\
            
            \bottomrule
        \end{tabular}
    }

    \caption{
    \textbf{Detailed Statistical Analysis.} 
    Performance reporting raw completion counts alongside percentages and 95\% bootstrap confidence intervals across different models.
    The best mean success rates are highlighted in \textbf{bold}.
    }
    \label{tab:confidence_intervals}
\end{table*}